\documentclass[pdflatex,sn-mathphys-num]{sn-jnl}

\usepackage{graphicx}%
\usepackage{subcaption}
\usepackage{amsmath,amssymb,amsfonts}%
\usepackage{amsthm}%
\usepackage{mathrsfs}%
\usepackage{xcolor}%
\usepackage{textcomp}%
\usepackage{manyfoot}%
\usepackage[utf8]{inputenc}%
\usepackage[T5]{fontenc}%
\usepackage{makecell}%
\usepackage{multicol}
\usepackage{supertabular}
\usepackage{makecell}
\usepackage{multirow}%
\usepackage{lipsum}
\usepackage{booktabs}
\usepackage{soul}
\usepackage{academicons}
\usepackage{float}
\usepackage{pifont}

\renewenvironment{table}[1][]%
{\tableorg[#1]%
\tablebodyfont%
\renewcommand\footnotetext[2][]{{\removelastskip\vskip3pt%
\let\tablebodyfont\tablefootnotefont%
\hskip0pt\if!##1!\else{\smash{$^{##1}$}}\fi##2\par}}%
}{\endtableorg}

\theoremstyle{thmstyleone}%

\theoremstyle{thmstyletwo}%

\theoremstyle{thmstylethree}%

\begin{document}

\title[Article Title]{ViTHSD: Exploiting Hatred by Targets for Hate Speech Detection on Vietnamese Social Media Texts}

\author[1,2,†]{Cuong Nhat Vo}\email{19520440@gm.uit.edu.vn}
\author[1,2,‡]{Khanh Bao Huynh}\email{19520625@gm.uit.edu.vn}
\author*[1,2,§]{Son T. Luu}\email{sonlt@uit.edu.vn}
\author[1,2,¶]{Trong-Hop Do}\email{hopdt@uit.edu.vn}

\affil[1]{\orgname{University of Information Technology}, \orgaddress{\city{Ho Chi Minh City}, \country{Vietnam}}}

\affil[2]{\orgname{Vietnam National University}, \orgaddress{\city{Ho Chi Minh City}, \country{Vietnam}}}

\affil[†]{\url{https://scholar.google.com/citations?user=-h52V_MAAAAJ&hl=en}}
\affil[‡]{\url{https://scholar.google.com/citations?hl=en&user=6TrJNjEAAAAJ}}
\affil[§]{\url{https://scholar.google.com/citations?user=wADRahMAAAAJ&hl=en}}
\affil[¶]{\url{https://scholar.google.com/citations?user=x4fM0EIAAAAJ&hl=en}}

\abstract{The growth of social networks makes toxic content spread rapidly. Hate speech detection is a task to help decrease the number of harmful comments. With the diversity in the hate speech created by users, it is necessary to interpret the hate speech besides detecting it. Hence, we propose a methodology to construct a system for targeted hate speech detection from online streaming texts from social media. We first introduce the ViTHSD - a targeted hate speech detection dataset for Vietnamese Social Media Texts. The dataset contains 10K comments, each comment is labeled to specific targets with three levels: clean, offensive, and hate. There are 5 targets in the dataset, and each target is labeled with the corresponding level manually by humans with strict annotation guidelines. The inter-annotator agreement obtained from the dataset is 0.45 by Cohen's Kappa index, which is indicated as a moderate level. Then, we construct a baseline for this task by combining the Bi-GRU-LSTM-CNN with the pre-trained language model to leverage the power of text representation of BERTology. Finally, we suggest a methodology to integrate the baseline model for targeted hate speech detection into the online streaming system for practical application in preventing hateful and offensive content on social media.}




\maketitle

\section{Introduction}
\label{intro}
Toxic content and harmful speech are now very popular on the Internet. With the growth of social network users, toxic content will continue to spread rapidly. There are several definitions of detrimental content on social networks such as hate speech, deceptive news, cyberbullying, abusive and toxic language, and sarcasm \cite{gongane2022detection}. To decrease the number of toxic comments and harassment posts, social network moderators need help from computers by using them as automatic detectors to identify which content is toxic. Fortuna and Nunes \cite{10.1145/3232676} defined hate speech as violence that attacks and expresses hatred toward a group of people based on their ethnic origin, nationality, religion, and gender even in subtle or rumor form. In this paper, we use the hate speech detection term because the comments in our research show individual hostility toward others. 

There are several attempts at the hate speech problem for Vietnamese social networks, such as the VLSP-HSD Shared Task 2019 \cite{vu2020hsd}, the ViHSD \cite{luu2021large} and the ViCTSD \cite{10.1007/978-3-030-79457-6_49} datasets. These works are valuable in constructing and evaluating the computer machine learning model to automatically identify toxic comments on social media space. Besides, the author in \cite{10.1007/978-3-030-97610-1_25} introduces a solution for hate speech detection on social networks by the streaming approach using the online streaming platform, which helps the system process for real-time processing. However, the research just focuses on the classification task, which means the output that is hate or not hate is based on the whole comments. According to \cite{kwarteng2022misogynoir}, it is difficult to capture the abuse and hateful content because users tend to replace the harmful words with more benign phrases and use subtle forms to prevent the detection. Thus, the challenge for the hate speech detection model is to exploit the hateful content, including not only the hateful action represented in the text but also the aimed target to increase the accuracy of classification models. In addition, measuring the public attitudes toward a specific target through the comments from social network media will help to understand more about the social phenomenon, and help solve the issues in social science \cite{chen2022analyzing,kim2024impact}. Moreover, from the output people not only want to find whether these comments are hate speech or not but also want to understand the specific targets that hateful comments are aimed at. According to \cite{MILLER20191}, this is called interpretability, which means "humans can understand the cause of a decision". The interpretability of the specific target in the hate speech can help mitigate the bias features in the users' text for hate speech detection \cite{gongane2024survey}. Therefore, in this paper, we provide new datasets named ViTHSD that help to interpret hate speech in the comments. 

Inheriting from the hate speech detection definition from \cite{vu2020hsd}, hate speech contains two main components: the target and the hate content aimed at the target. Our new dataset is focused on the targets in each comment, and each target will have a level to measure the hateful meaning. Thus, the task of identifying the target hate speech detection is denoted as below:
\begin{itemize}
    \item \textbf{Input:} A comments from social network.
    \item \textbf{Output:} A list contains the terms $target\#level$, where $target$ indicates the target mentioned in the comments, and the $level$ determines the level of hate. 
\end{itemize}

From the definition above, the targets used in our new dataset include individuals, groups, religions, and politics. There are three levels: clean, offensive, and hate. The meaning and explanation of those targets and levels are shown in Section \ref{dataset}. We called this task the targeted hate speech detection task, which is similar to the aspect-based sentiment analysis (ABSA). To solve this task, we first construct a new dataset for targeted hate speech detection on Vietnamese texts. We use the available text comments data from the ViHSD dataset \cite{luu2021large} and manually annotate the new labels with detailed annotation guidelines and strict annotation procedures. After the data creation process, we have a dataset with 10K comments and 5 different targets with 4 levels of hate speech. Then, we use the multi-label text classification approach to solve the task. To build the model for identification targets and their level of hate speech, we use the Bi-GRU-LSTM-CNN \cite{van2019hate} model, which achieves good performance in the VLSP2019 share task about Hate speech detection \cite{vu2020hsd}. Besides, to leverage the power of BERTology models in text representation, we proposed a baseline that combines the BERTology models with the Bi-GRU-LSTM-CNN to enhance the accuracy of the model for the targeted hate speech detection task. To evaluate the performance between models, we separate the main task into two small tasks: only target detection and the target with level detection and evaluation with the Precision, Recall, and F1-score metrics \cite{Nguyen_Nguyen_Ngo_Vu_Tran_Ngo_Le_2019}. Finally, we suggest a methodology to integrate the baseline model for targeted hate speech detection into the online streaming system for detecting hateful comments from social platforms by leveraging real-time data processing tools such as Spark Streaming and Apache Kafka for processing real-time comments coming from live stream videos on Youtube.  

Overall, our main contributions to this paper are:
\begin{itemize}
    \item First, we introduce a new dataset for targeted hate speech detection on Vietnamese social media texts with 10K human-annotated comments to train and evaluate the model. The dataset is manually annotated with strict guidelines and explicit procedures to ensure objectivity in labels. 
    \item Second, we propose a baseline model for targeted hate speech detection on social media texts based on the multi-label text classification task. We choose the BERTology approach \cite{rogers-etal-2020-primer} because they perform well on the social text according to the benchmark from \cite{nguyen2022smtce}.
    \item Third, we propose a solution to process real-time comments from social live-stream platforms. We also test our proposed system on a real-time streaming video to evaluate its efficiency and show potential applications for future usage. 
\end{itemize}

Our paper is structured as follows. Section \ref{relatedworks} shows previous works and attempts on the hate speech detection problems in Vietnamese and other low-resource languages. Section \ref{dataset} introduces our dataset, including the data creation process, data annotation guidelines, and data statistical information. Next, Section \ref{method} illustrates our proposed method to build the classification model for the targeted hate speech detection task and the real-time comments processing techniques. Then, Section \ref{results} shows our empirical results and error analysis to find the advantages and disadvantages of the current model. Section \ref{result_online_stream} describes the execution evaluation and practical results of the targeted hate speech detection model in the online streaming system. Finally, Section \ref{conclusion} concludes our works and proposes further research. 
\section{Related Works}
\label{relatedworks}
In Vietnamese, the VLSP-HSD comes from the VLSP2019 Shared Task, the earliest work for Vietnamese Hate Speech Detection. Each comment in the VLSP-HSD has one of three labels: CLEAN, OFFENSIVE, and HATE. Following this work, Luu et al. \cite{luu2021large} increase the number of comments to construct a large-scale dataset with 33,400 comments and also provide a strict annotation scheme to assure the quality of labeled data. In addition, the ViHOS \cite{hoang2023vihos} is a new dataset that can identify the position of hate speech in the comments by character. This dataset opens a new task that extracts hateful terms from comments of users instead of just classifying them as hate or not. In general, those datasets have not focused on the targets aimed in the comments. Hence, our new dataset - the ViTHSD concerned about what is mentioned in the comments and the level of hatefulness aimed at. Overall, Table \ref{table:table_1} shows the comparison between available datasets for Vietnamese Hate Speech Detection. Besides those mentioned datasets in Table \ref{table:table_1}, we would like to introduce other available datasets for Vietnamese toxic and cyberbullying detection on social media sites. For example, the UIT-ViCTSD \cite{10.1007/978-3-030-79457-6_49} is a large-scale dataset used for the constructive and toxic speech detection on Vietnamese News texts, and the UIT-ViOCD \cite{nguyen2021vietnamese} dataset is used for complaint detection on E-commerce websites. Apart from Vietnamese, several datasets are used for targeted hate speech detection in other low-resource languages, such as the dataset provided by Ousidhoum et al. \cite{ousidhoum-etal-2019-multilingual} and Albadi et al. \cite{albadi2019investigating} on the Abrabic, the L-HSAB dataset \cite{mulki-etal-2019-l} and the dataset provided by AbdelHamid et al. \cite{abdelhamid2022levantine} for Levantine, the dataset provided by Ibrohim and Burdi \cite{ibrohim-budi-2019-multi} on Indonesian, and the Hatexplain dataset \cite{mathew2021hatexplain} for English. From those datasets, it can be seen that the inter-annotator agreement score is just in the moderation score. For example, the Hatexplain dataset \cite{mathew2021hatexplain} obtained 0.46 by Krippendorff $alpha$ and the dataset by Ousidhoum et al. \cite{ousidhoum-etal-2019-multilingual} obtained only 0.202 by Krippendorff $alpha$. This indicates the challenge in the annotation procedure when constructing the dataset.

\begin{table}[ht]
    \centering
    \caption{Comparison of different hate speech datasets for the Vietnamese language}
    \label{table:table_1}
    \resizebox{\textwidth}{!}{
    \begin{tabular}{|l|p{4cm}|c|p{3.1cm}|p{1cm}|p{1.5cm}|}
        \hline
        \textbf{Dataset} & \textbf{Labels} & \textbf{Size} & \textbf{Task} & \textbf{Target Labels} & \textbf{Rationales} \\ [0.5ex]
        \hline
        VLSP-HSD (2019) \cite{vu2020hsd} & Hate, Offensive, Normal & 20,345 & Multiclass classification & No & No \\
        \hline
        ViHSD (2021) \cite{luu2021large} & Hate, Offensive, Normal & 33,400 & Multiclass classification & No & No \\
        \hline
        ViCTSD \cite{10.1007/978-3-030-79457-6_49} & Toxic and Non-toxic & 10,000 & Binary classification & No & No \\
        \hline
        ViHOS (2023) \cite{hoang2023vihos} & Spans determining the hate speech pieces in the comments & 11,065 & Span labeling & No & Yes \\
        \hline
        \textbf{ViTHSD (Ours)} & Hate, Offensive, Normal corresponding to each of five targets & 10,000 & Multilabel classification & Yes & No \\
        \hline
        \end{tabular}
        }
\end{table}

Luan et al. \cite{nguyen2022smtce} take a particular investigation about the SOTA methods applied to Vietnamese social text, including monolingual and multilingual BERTology. This work proposed SMCTE - a benchmark evaluation framework. However, SMCTE can apply for the multi-class classification task only, which cannot be suitable for our task on targeted detection by multi-label classification. However, the configurations of BERTology models used in this project are helpful for our methods. Besides, the Bi-GRU-LSTM-CNN \cite{van2019hate} achieved the highest results for the hate speech detection and job prediction tasks \cite{9140760}. This is an ensemble model that combines three different layers including the convolutional, the Bi-LSTM, and Bi-GRU to enhance the feature extraction module for natural language text. However, this approach is used for the multi-class classification task, which cannot directly apply to our problem. Instead, the BERT model fine-tuned for the Aspect-based sentiment analysis (ABSA) task proposed by Dang et al. \cite{9865479} is appropriate for our task. Therefore, we fine-tune the BERT architecture based on the work in \cite{9865479} with different BERTology models as shown in \cite{nguyen2022smtce} for the targeted hate speech detection task, and combine BERTology with the Bi-GRU-LSTM-CNN to construct a model for automatically detecting targets and hateful level in human texts. 

Finally, Khanh et al. \cite{quoc2023vietnamese} proposed an online streaming system for detecting hate speech comments with efficient data-processing steps to improve the performance of the classification models based on the ViHSD dataset \cite{luu2021large}. Besides,  Vo et al. \cite{10.1007/978-3-030-97610-1_25} and Doan et al. \cite{9994299} proposed a real-time streaming system on social platforms in Vietnam such as Facebook and YouTube to detect hateful online comments. However, these two works are applied based on the ViHSD \cite{luu2021large} and the VLSP-HSD \cite{vu2020hsd} datasets, which cannot be applied to the targeted hate speech problem. Hence, based on the system architecture proposed by two previous works, we construct a new system that can treat the online streaming comments and automatically detect hate speech comments at the target level. 

\section{The dataset}
The primary goal of this section is to offer comprehensive insights into the different strategies that we have implemented to annotate the dataset, the data that we have relied upon to enable meaningful analysis, and the evaluation methodology that we have employed to ensure the validity and reliability of our findings. Additionally, we will provide a detailed exposition of the evaluation methodology utilized to assess the accuracy and robustness of our detection system. This rigorous evaluation process would facilitate our models' refinement and improve the system's performance, leading to a higher degree of efficacy in detecting instances of hate speech.
\label{dataset}

\subsection{Annotation Procedure}
\textbf{Data source}: Our dataset is derived from the ViHSD data, a study that aimed to identify hate speech on Vietnamese social media. The Vietnamese Hate Speech Detection dataset (ViHSD) \cite{luu2021large} is a human-annotated dataset developed to automatically detect hate speech on social networks. This dataset comprises over 30,000 comments, each labeled as CLEAN, OFFENSIVE, or HATE. We removed comments that contained links, images, or videos that did not provide any information, selecting over 10,000 comments for the annotation process. However, we retained emoticons in the text since they could contain essential information for annotators.

\textbf{Targets in Hate speech detection}: According to  \cite{10.1145/3232676}, although the hate speech definition mentions various aspects to detect the hate speech, the general hate speech on social media focuses on some specific targets such as Sexual orientation, Race, and Religion. Other targets such as Gender and Disability are very few thus they have less meaning when separated. Besides, the authors in \cite{bonaldi-etal-2024-nlp} show various types of hate according to different aspects such as culture, economics, crimes, rapism, women oppression, history, and other/generic types. However, some hate speech appears in complex linguistic styles (e.g. indirect sarcasm and humor), making it hard to mitigate the bias among targets. Referred from the previous works in \cite{10.1145/3232676,mathew2021hatexplain,schmidt-wiegand-2017-survey,ER}, we proposed a list of targets that mostly appear in Vietnamese comments including \textit{Groups, Individuals, Religion, Race, and Politics}. These targets play the role of exploiting the subject expressed in the comments by users, which helps efficiently detect hate speech by targets. The details about these targets are described in the Annotation guidelines section.  

\textbf{Annotation guidelines}: The ViTHSD dataset comprises four hatred labels, namely HATE, OFFENSIVE, CLEAN, and NORMAL, pertaining to five different targets. Even though a comment may contain multiple targets in ViTHSD, the degree of hate or abuse involved can vary for each target. Table \ref{Annotation_guidelines} describes the definitions and examples of each hatred level. There are three hatred levels, which are CLEAN, OFFENSIVE, and HATE. The NORMAL is not the hatred level since it indicates that the target is not mentioned. The principal objective of the annotation process is to identify the groups that are being targeted by hate speech in the comments and determine the extent of hostility directed towards those communities. To ensure that the labeling process is facilitated, we decided to base the selection of target groups on the characteristics of the dataset. We included five targets based on individuals (1), groups (2), religion/creed (3), race/ethnicity (4), and politics (5). The details of those targets are described below: 

\begin{itemize}
\item \textbf{Individuals} which is Hate speech based on the identities of individuals can be attacked on elements of physical disability, sexual orientation, etc. This is one of the most popular attack methods on Vietnamese social networks, often expressed in sentences containing names or personal pronouns in the Vietnamese language. According to a report by Vietnam Digital 2021, 53\% of Vietnamese internet users have experienced online harassment, with 21\% of them experiencing hate speech \cite{luu2021large}. These types of personal attacks are commonly referred to as cyberbullying and cyberaggression \cite{chatzakou2019detecting}. For example: \\

\emph{Comment 1:} Thứ \underline{ngu} như \underline{m} nói chuyện mất thời gian...đi học lại mẫu giáo đi  (English: Talking to a \underline{stupid} like \underline{you} is wasting time.). Comment 1 contains the phrase \emph{stupid}, which means \emph{ngu} referring to an individual as having an intellectual disability is considered an insult to persons who may have a cognitive or intellectual disability. The term "\emph{ngu}" is often used in this context to imply that the individual is deficient in some way and, as a result, is inferior to others in terms of cognitive abilities or intelligence. The use of such pejorative language can be highly offensive and hurtful toward people with disabilities, and it is frequently regarded as a form of hate speech that reinforces damaging stereotypes and biases directed toward individuals based on their perceived abilities and differences. \\

\item \textbf{Groups} which is Hate speech can target groups or organizations, along with personal attacks. This is one of the most common forms of attack on social networks in Vietnam and one of the most common types of content. As discussed by Parikh et al. \cite{parikh2021categorizing}, hate speech commonly takes the form of discriminatory attacks and disregard towards women. \textbf{Sexism} and \textbf{misogyny} are two introduced concepts, where Sexism refers to discrimination based on gender that mainly affects women, while misogyny refers to hatred or deep hatred for women. Non-governmental organizations (NGOs) are often the target of hate speech. For example: \\

\emph{Comment 2:} \underline{Đm chúng mày}. (English: \underline {F*cking you guys}). Comment 2 includes the term \emph{chúng mày}, a compound word commonly used in Vietnamese to refer to a group of people as equal or inferior, often used to display either contempt or intimacy. When used in conjunction with the phrase \emph{Đm}, it signifies disdain in this particular context. \\

\item \textbf{Religion/creed} which is hate speech can target individuals based on their religious affiliation or beliefs. This is often referred to as "religious hate speech". In addition, hate speech can also target individuals according to their creed, which includes their beliefs or faith, and this is also considered a form of hate speech that promotes intolerance and divisive attitudes \cite{ibrohim2019multi}. For example: \\

\emph{Comment 3:} \underline{tiên tri xàm}. (English: \underline {Prophet speaks nonsense}). Comment 3 contains the phrase \emph{speaks nonsense}, which means \emph {xàm}. By using this phrase, the commenter is effectively dismissing the prophet's teachings as baseless or irrational, suggesting a lack of respect for the religious views held by the prophet and their followers. Such comments can be seen as a form of hate speech that seeks to belittle and denigrate individuals based on their religious beliefs and can be harmful and offensive to those targeted. \\

\item \textbf{Race/ethnicity} which is hate speech targets individuals or groups based on their race or ethnicity. It's a form of discriminatory behavior that can be very hurtful and offensive. This type of hate speech is often used to marginalize and oppress individuals based on their physical characteristics, such as the color of their skin, their facial features, their height, and other characteristics. Through the use of derogatory language and the spread of negative stereotypes, hate speech based on race perpetuates harmful attitudes toward individuals and particular communities, further reinforcing harmful biases and prejudices \cite{macavaney2019hate}. For example: \\

\emph{Comment 4:} Đạo khổng ăn quá sâu vào tiềm thức rồi. Xã hội đó không thể bắt kịp nền văn minh nhân loại nữa. (English: Taoism is far too deeply embedded in the subconscious. That society was no longer capable of keeping up with human civilization.). Comment 4 contains a biased statement that is directed at \textit{Taoism} and suggests that this religion has had a deep and lasting impact on the subconscious mind of the speaker. The speaker also implies that society at large has not kept pace with the development of human civilization because of the influence of this religion. Such comments can be considered a form of prejudice or discrimination because they seek to denigrate or belittle a particular religion, culture, or community without providing substantial or credible evidence. \\

\item \textbf{Politics} which is hate speech is centered on political factors, including attacking governmental institutions, heads of state, and other related entities. Such speech is not only deeply hurtful and divisive but can have serious and far-reaching consequences for individuals, communities, and nations. For example: \\

\emph{Comment 5:} DM \underline{ lũ tham quan} bán nước. (English: F*ck \underline {the corrupt officials} who sell the country.). In Comment 5, there is an offensive sentence that uses the word f*ck and includes the phrase \textit{lũ tham quan}, which means \textit{the corrupt officials} about high-ranking government officials who engage in unethical practices. This statement conveys a strong sense of hostility towards these officials. \\
\end{itemize}

\begin{table}[ht!]
    \centering 
    \caption{Definition of hatred level and examples}
    \label{Annotation_guidelines}
    \resizebox{\textwidth}{!}{
        \begin{tabular}{|c|c|c|}
        \hline
        \textbf{Label} & \textbf{Description} & \textbf{Example}  \\ 
        \hline
        NORMAL (*) & \multicolumn{1}{m{5cm}|}{Comments with content that is not related to the targets under consideration. There is no need to consider the comments containing offensive words, hate or not.}  & \multicolumn{1}{m{5cm}|} {\textbf{Example 1:} Nhanh thực sự (English: It is really fast). 
        This comment is thoroughly clean, contains no offensive or hate words, and attacks no one. } \\
        \hline
        CLEAN & \multicolumn{1}{m{5cm}|} {Comments should pertain to the discussed targets, without offensive or hateful language. Additionally, comments devoid of intention to insult the mentioned targets shall not be included.} & \multicolumn{1}{m{5cm}|} {\textbf{Example 2:} <person name> Mua được mấy cái siêu xe cho các con du học bên bển rồi:v (English: <person name> Bought some great cars for your kids to study abroad:v). 
        This comment contains the noun "<person name>". However, it does not contain any word that is offensive or hateful. } \\
        \hline
        OFFENSIVE & \multicolumn{1}{m{5cm}|} {Comments have content related to the targets considered. Include words that are offensive or hateful, but content that does not have the intent to insult the targets mentioned in the comments.} & \multicolumn{1}{m{5cm}|} {\textbf{Example 3:} Dm cạn lời với bọn này (English: F*ck, I'm at a loss for words with these people.).
        This comment contains the phrase \textit{F*ck}. However, its content is not intended to offend or be offensive to a specific target audience. } \\
        \hline
        HATE & \multicolumn{1}{m{5cm}|} {Comments have content related to the targets considered. Some exceptional cases happened with the HATE label:\newline 
        \textbf{Case 1: }Include words that are offensive or hateful, content that has the intent to offend the targets mentioned in the comments.\newline
        \textbf{Case 2: }Do not include words that are offensive or hateful, but content that is intended to offend the targets mentioned in the comments. Often this content is of a figurative or metaphorical nature. It is difficult to identify these comments, as annotators are required to have broad knowledge and understanding in many areas.} & \multicolumn{1}{m{5cm}|} {\textbf{Example 4:} Đi tìm nó đánh chết mẹ nó luôn chị (English: Go find him and kill his mother too, sister.).
        The phrase \textit{kill his mother} appears in this comment. This is an inflammatory phrase that can cause hatred and has the potential to threaten to cause anyone to feel unsafe. \newline
        \textbf{Example 5:} Bán Nam bán Nữ (English: semi-male, semi-female.).
        No words expressing sarcasm or hatred, but content implies insulting targets, specifically people's sexual characteristics.} 
        \\
        \hline
    \end{tabular}
    }
   \textit{(*) NORMAL is not the hatred level because it expresses that the target is not mentioned. There are three hatred levels including CLEAN, OFFENSIVE, and HATE.}
\end{table}

\textbf{Criteria for annotators}: After collecting data and developing an initial annotation procedure, the next step was to recruit annotators. In this research, the criteria for the selection of annotators are as follows:  All of the annotators are native speakers of Vietnamese. They do not belong to any political party/organization. The annotators are from different regions and they have different religious beliefs. As the annotation of hate speech is highly subjective, this is done to reduce bias \cite{ibrohim2019multi}. Based on these criteria, a group of students was hired for the annotation tasks, with a gender split of 0.25\% women and 0.75\% men.

\begin{figure}[H]
    \centering
  \includegraphics[width=\textwidth]{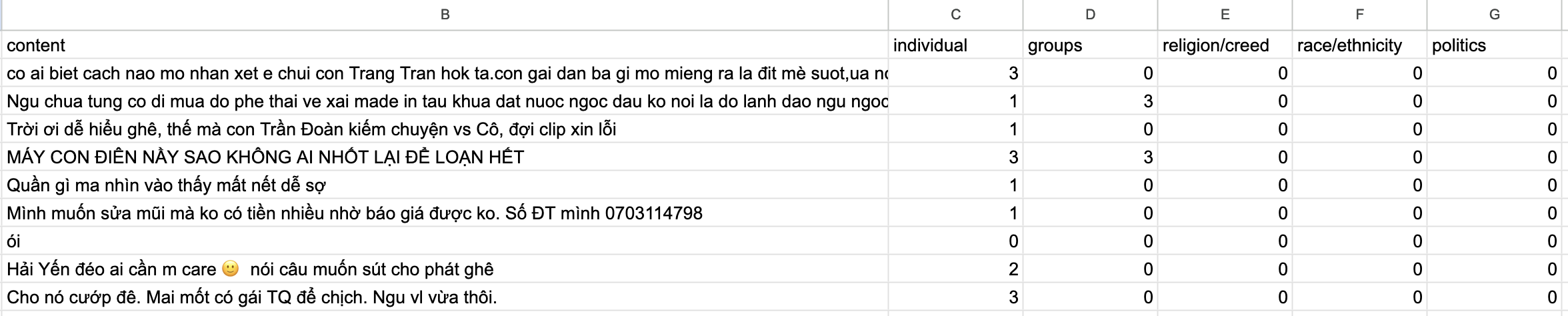}
    \caption{An example of annotated data by users on Google sheet.}
    \label{fig:annotationexample}
\end{figure}

\textbf{Annotation tool}: We used the Google sheet\footnote{\url{https://docs.google.com/spreadsheets}} for annotating the data. Each annotator annotates their label on their own sheet as shown in Figure \ref{fig:annotationexample}. Each comment (one comment per row) has five columns corresponding to five targets. The annotator assigns one numeric value denoting the hatred level for each target as 1 for clean, 2 for offensive, and 3 for hate (see Table \ref{Annotation_guidelines} for details about the level of hatred level). If the comment does not mention the target, the annotator assigns the value 0 as the normal label (as shown in Table \ref{Annotation_guidelines}).

\subsection{Dataset Creation Steps}
\begin{figure}[ht]
  \centering
  \includegraphics[width=\textwidth]{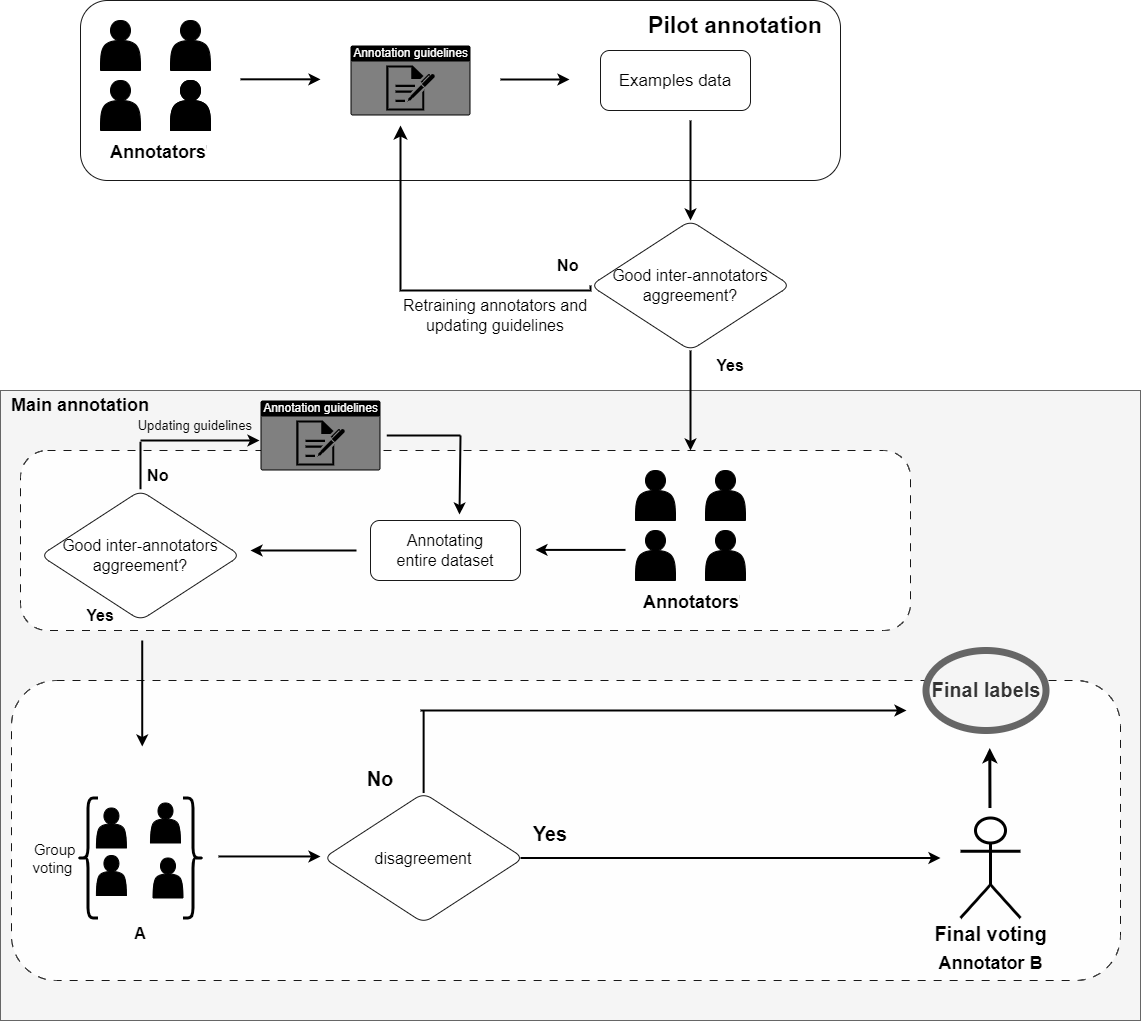}
  \caption{Data annotation process for the ViTHSD dataset with $k$ > 0.4 is acceptable \cite{landis1977measurement}.}
  \label{fig: Data annotation }
\end{figure}

The dataset creation process began with a pilot annotation round, followed by the main annotation task. The main annotation task was inspired by the work of Mathew et al. \cite{mathew2021hatexplain} and Luu et al. \cite{luu2021large}. Inter-annotator agreement (IAA) was computed using the Cohen Kappa index ($k$) \cite{cohen1960coefficient}, and the final labels were selected using a majority voting algorithm \cite{parhami1994voting}. Figure \ref{fig: confusion_matrix_10K} displays the inter-annotator agreement between each annotating pair for each target. The dataset was generated using a two-stage process, as depicted in Figure \ref{fig: Data annotation }.\newline

\textbf{Pilot annotation}: The pilot annotation phase gave annotators 100 comments to practice and become familiar with the task before the main annotation phase. Annotators needed to identify the target community and classify the level of hate or abuse for each aspect of the comments. Since a comment could have multiple targets, but different levels of hate/abuse for each one, examples and explanations were given for novice annotators to understand the labeling process better. The agreement between annotators was assessed using Landis \cite{landis1977measurement}, with $k$ > 0.4 considered moderate. If $k$ < 0.4, we sought feedback from annotators on the annotation process to enhance our approach and update instructions for the main annotation task. 

\textbf{Main annotation}: After completing the pilot annotation phase, we carefully evaluated the quality of our annotations and continued with the main annotation task. We divided into several rounds for classification and evaluation, in each round we will provide the annotator with about 1000 comments. Following the labeling process, we evaluated the level of agreement between the annotators to gain insights into their collective comprehension and thought processes. The final label is decided by a majority vote of the annotators if $k$ > 0.4, and the annotators engage in a thoughtful discussion to improve the labeling process and update guidelines to cultivate the best dataset if $k$ < 0.4. Our final dataset comprises 10,000 comments, and we have observed a moderate agreement achieved through the Cohen-Kappa K-average metric ($k$ = 0.45) for agreement among the annotators in the dataset \cite{landis1977measurement}.

Using Landis criteria \cite{landis1977measurement} to calculate inter-annotator agreement is a widely accepted practice in natural language processing and other fields that involve manual annotation. The Landis criteria, Cohen's kappa coefficient \cite{cohen1960coefficient} in particular, provide a reliable and valid measure of inter-annotator agreement. As stated in the original statement, a threshold of $k$ > 0.4 is considered moderately acceptable. If the inter-annotator agreement falls below the threshold of $k$ < 0.4, feedback from the annotators may be necessary to identify and correct issues in the annotation process. Possible areas of concern include unclear instructions, inadequate annotation standards, differing interpretations of data, or insufficient training for annotators. In cases where there is no perfect inter-annotator agreement, a consensus approach to determining the final label is often necessary. Majority voting is suggested as a reliable method for determining the final label, particularly in situations where multiple raters are involved, or when the raters are not subject matter experts \cite{bobicev2017inter}.

\subsection{Inter-annotator agreement}

\begin{figure}[ht]
\centering
  \includegraphics[width=.9\textwidth]{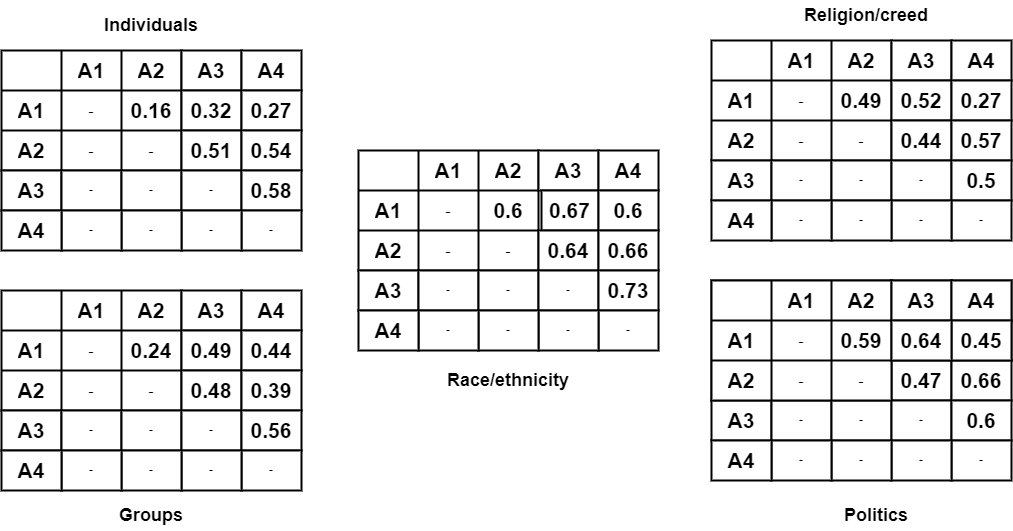}
    \caption{The confusion matrix between the annotators in a set of 5,000 comments in each target without levels, computed by the Cohen kappa index  ($k$)}
    \label{fig:confusion_matrix_without_level5K}
\end{figure}

Cohen's kappa ($K$) \cite{cohen1960coefficient}, Fleiss' kappa ($K$) \cite{fleiss1971measuring}, and Krippendorf's alpha \cite{artstein2008inter} are commonly used metrics for measuring inter-annotator agreement. Cohen's kappa and Fleiss' kappa are Chance-Corrected Coefficients that evaluate agreement beyond the chance level. These metrics provide a numerical measurement of agreement between annotators and help assess annotation reliability \cite{bobicev2017inter}. Fleiss' kappa is preferred for more than two annotators since Cohen's kappa only applies to two annotators. For the study, Cohen's kappa was employed and calculated for every annotator pair on each target before being averaged across the entire dataset. Moreover, to evaluate the level of difficulty of the dataset, we computed the agreement among targets without levels to compare targets with levels.

\begin{figure}[ht]
\centering
  \includegraphics[width=.9\textwidth]{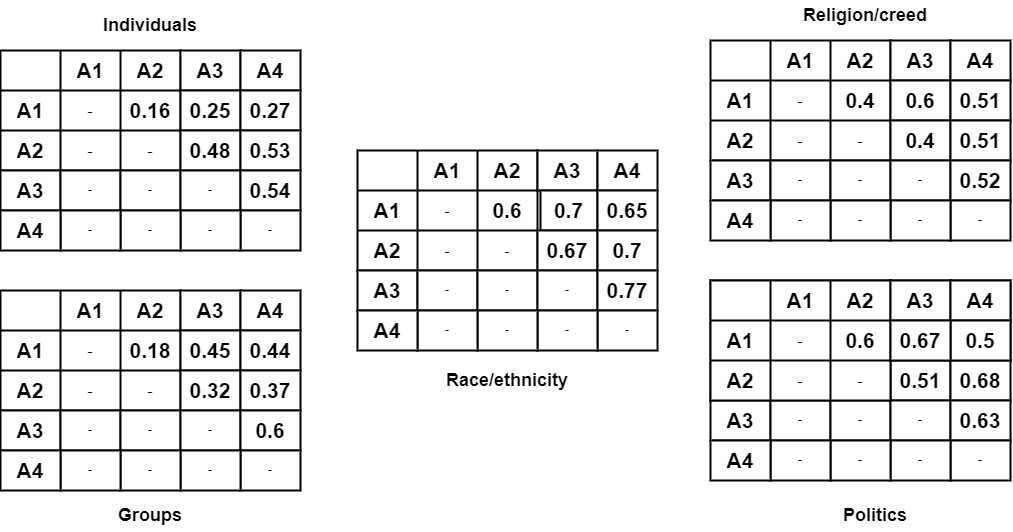}
    \caption{The Cohen's kappa index ($k$) was used to compute the confusion matrix between annotators in a set of 10,000 comments for each target without levels.}
    \label{fig:confusion_matrix_without_level10K}
\end{figure}

Figures \ref{fig:confusion_matrix_without_level5K} and Figures \ref{fig:confusion_matrix_without_level10K} demonstrate the confusion matrix for comments between each pair of annotators. Table \ref{Average_Kappa_without_levels} exhibits the average agreement level attained by annotators for each goal and the complete dataset. The annotators had a high agreement on the \textit{individuals} and \textit{Politics} targets, while \textit{Religion/Creed}  received lower scores. The overall inter-annotator agreement Cohen's Kappa average was 0.5 and 0.51 for the 5,000 and 10,000 datasets, respectively.

\begin{table}[ht]
    \caption{Cohen's Kappa ($k$) average for each aspect without levels.}
    \label{Average_Kappa_without_levels}
    \begin{tabular}{|c|ccccc|c|}
    \hline
    \multirow{2}{*}{\# Comments} & \multicolumn{5}{c|}{Target}                                                    & \multirow{2}{*}{$k$} \\ 
    \cmidrule{2-6}
    & \multicolumn{1}{c|}{individuals} & \multicolumn{1}{c|}{ Group} & \multicolumn{1}{c|}{Religion/creed} &\multicolumn{1}{c|}{Politics} &\multicolumn{1}{c|}{Race/ethnicity}  &                   \\ \hline
    5,000  & \multicolumn{1}{c|}{0.4} & \multicolumn{1}{c|}{0.43} & \multicolumn{1}{c|}{0.47} & \multicolumn{1}{c|}{0.57} &\multicolumn{1}{c|}{0.65} &   0.5                \\ \hline
    10,000  & \multicolumn{1}{c|}{0.37} & \multicolumn{1}{c|}{0.4} & \multicolumn{1}{c|}{0.49} & \multicolumn{1}{c|}{0.6} &
    \multicolumn{1}{c|}{0.68} &   0.51                \\ \hline
    \end{tabular}
\end{table}

\begin{table}[ht]
    \caption{Average Cohen's Kappa ($\overline{k}$) for each target with levels}
    \label{Average_Kappa}
    \begin{tabular}{|c|ccccc|c|}
    \hline
    \multirow{2}{*}{\# Comments} & \multicolumn{5}{c|}{Target}                                                    & \multirow{2}{*}{$\overline{k}$} \\ \cmidrule{2-6}
        & \multicolumn{1}{c|}{individuals} & \multicolumn{1}{c|}{ Group} & \multicolumn{1}{c|}{Religion/creed} &\multicolumn{1}{c|}{Politics} &\multicolumn{1}{c|}{Race/ethnicity}  &                   \\ \hline
        5,000  & \multicolumn{1}{c|}{0.41} & \multicolumn{1}{c|}{0.43} & \multicolumn{1}{c|}{0.33} & \multicolumn{1}{c|}{0.41} &
        \multicolumn{1}{c|}{0.48} &   0.41                \\ \hline
        10,000  & \multicolumn{1}{c|}{0.53} & \multicolumn{1}{c|}{0.41} & \multicolumn{1}{c|}{0.33} & \multicolumn{1}{c|}{0.57} &
        \multicolumn{1}{c|}{0.4} &   0.45             \\ \hline
    \end{tabular}
\end{table}

\begin{figure}[ht]
    \centering
  \includegraphics[width=.85\textwidth]{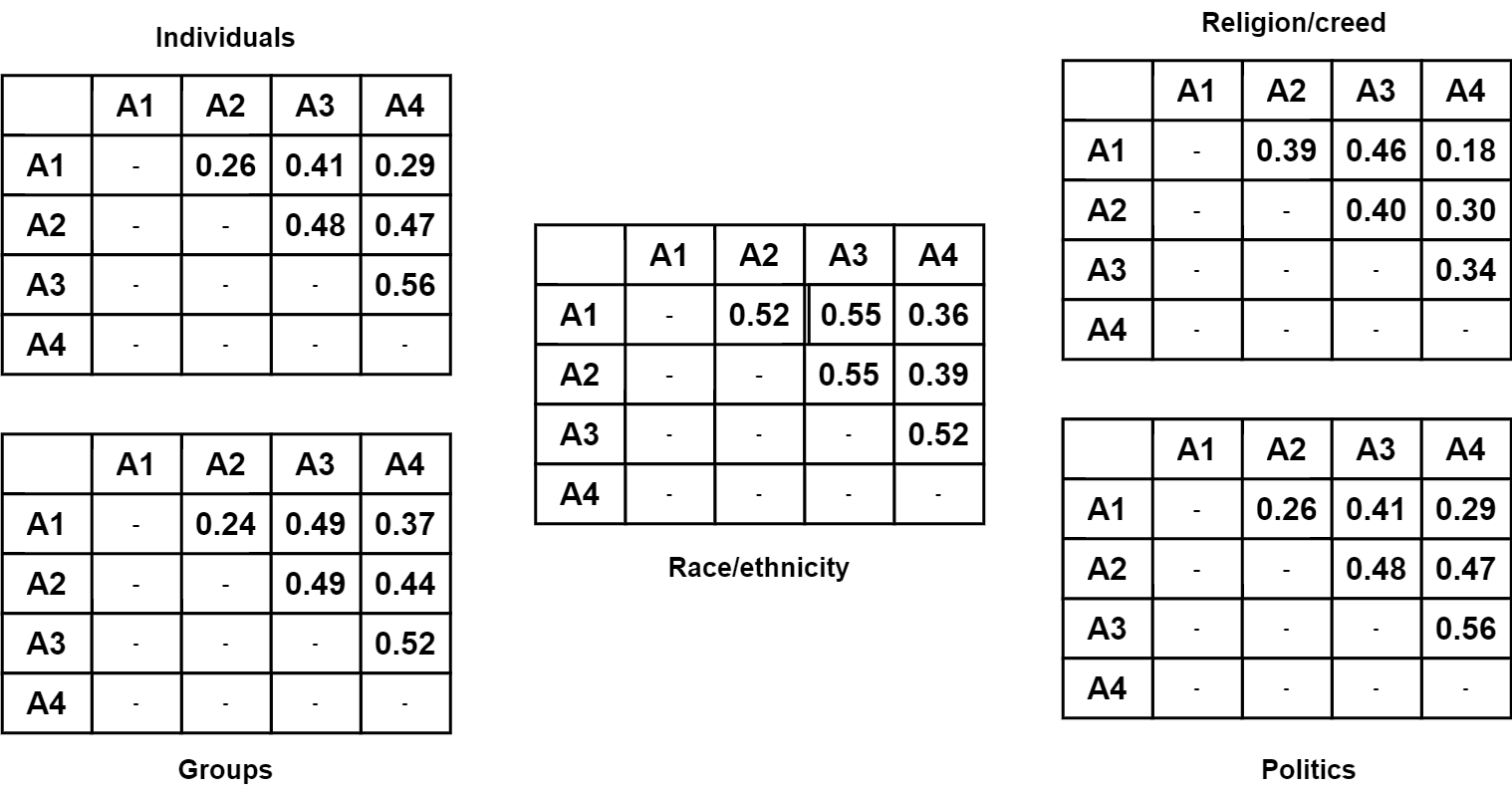}
    \caption{The Cohen's kappa index ($k$) was used to calculate the confusion matrix between the annotators on 5,000 comments per target.}
    \label{fig: confusion_matrix_5K}
\end{figure}

\begin{figure}[ht!]
\centering
  \includegraphics[width=.85\textwidth]{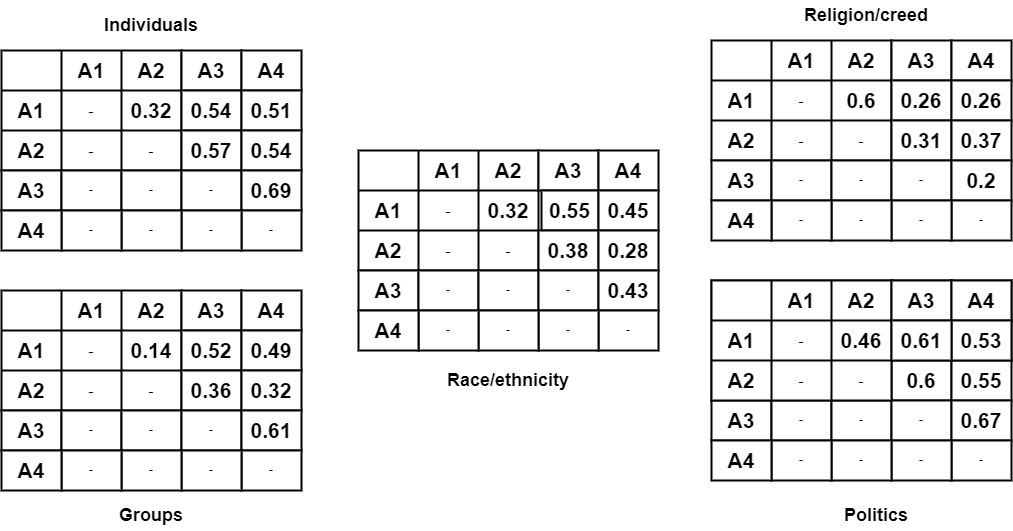}
    \caption{The Cohen's kappa index ($k$) was used to calculate the confusion matrix between the annotators on 10,000 comments per target.}
    \label{fig: confusion_matrix_10K}
\end{figure}

We found significant variation in the average Cohen's Kappa results for each aspect with levels. Table \ref{Average_Kappa} shows the inter-annotator agreement analysis results for a dataset of 5,000 comments, indicating moderate agreement among annotators. The \textit{Race/Ethnicity} target had the highest level of agreement, while \textit{Religion/Creed} had the lowest. We also analyzed a larger dataset of 10,000 comments, which showed higher agreement levels among annotators than the smaller dataset. However, for the larger dataset, \textit{individuals} and \textit{Religion/Creed} targets had lower agreement scores compared to \textit{Group} and \textit{Race/Ethnicity}. Figures \ref{fig: confusion_matrix_5K} and \ref{fig: confusion_matrix_10K} provide clearer consensus between annotator pairs.\\

Overall, analysis of the inter-annotator agreement shows that the annotators agree to a high degree when determining the aspect of a sentence. This suggests a shared understanding of the underlying themes commented upon. However, there was significant variability between annotators when assessing the level of hate within a given aspect. This variability was particularly evident for the \textit{Religion/Creed} and \textit{Race/Ethnicity} targets, suggesting that the perception of hate speech related to these two targets is more subjective and less consensual among annotators. These findings highlight the complexity of identifying and assessing hate speech, especially when the targets are related to sensitive issues such as ethnicity and religion.


\subsection{ Dataset Analysis}
We divided our dataset into the training, development, and test sets with a proportion of approximately 70/10/20. Table \ref{tb_overview} illustrates the general statistics about the ViTHSD dataset. The vocabulary size is calculated by token level, and the average length is calculated by the total length of comments divided by the number of comments in each set. From Table \ref{tb_overview}, it can be seen that the train set has the biggest vocabulary size. However, the development (dev) set has the highest average length value, which means that the comments in the development set seem longer than the train and dev sets. In contrast with the average length, the vocabulary size in the dev set is lower than the two remains. 

\begin{table}[H]
    \centering
    \captionsetup{width=\linewidth}
    \caption{Overview statistic about the ViTHSD dataset}
    \label{tb_overview}
    \begin{tabular}{|l|l|l|l|}
            \hline
             & \textbf{train} & \textbf{dev} & \textbf{test} \\ \hline
            \textbf{Num. comments} & 7,000          & 1,201        & 1,800         \\ \hline
            \textbf{Vocab. size}   & 12,701         & 4,547        & 5,684         \\ \hline
            \textbf{Avg. length}   & 57.33          & 58.23        & 55.54         \\ \hline
    \end{tabular}        
\end{table}

\begin{figure}[H]
    \centering
    \includegraphics[width=.7\textwidth]{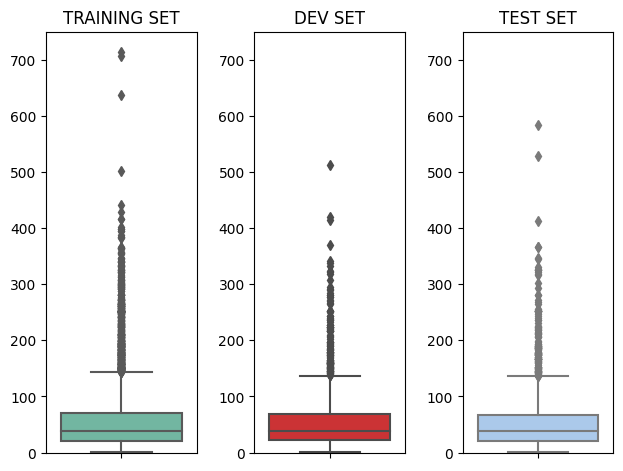}
    \caption{The distribution of length of comments on the training, development, and test sets of the ViTHSD}
    \label{fig_length_distribution}
\end{figure}

On the other hand, we also conduct a statistical analysis of the comments length distribution on the training, development, and test sets. Table \ref{tbl_length_dist_analysis} shows the statistical information about the length of comments on those three sets. Besides, Figure \ref{fig_length_distribution} illustrates the boxplot graph according to the information from Table \ref{tbl_length_dist_analysis}.

\begin{table}[H]
    \centering
    \captionsetup{width=\linewidth}
    \caption{Statistical information about the length distribution on the training, development, and test sets of the ViTHSD}
    \label{tbl_length_dist_analysis}
    \begin{tabular}{l|l|l|l}
        & \textbf{Train} & \textbf{Dev} & \textbf{Test} \\ \hline
        \textbf{Min}    & 1              & 1            & 1            \\ \hline
        \textbf{25\%}   & 21             & 22           & 21            \\ \hline
        \textbf{Median} & 38             & 38           & 38            \\ \hline
        \textbf{75\%}   & 70             & 68           & 67            \\ \hline
         \textbf{Max}    & 714            & 512          & 583   
    \\ \hline
    \end{tabular}
\end{table}

According to Figure \ref{fig_length_distribution}, the median length of comments in the three sets is the same, which is 38 according to Table \ref{tbl_length_dist_analysis}. Also, the training, development, and test sets have a minimum length of comments equal to 1. However, the maximum length of comments in the three sets is not the same. The training set has the highest maximum length of comments, which is 714 while the development set has the lowest maximum length of comments, which is 512. In general, the three sets of the ViTHSD have the same length distribution.

\begin{table}[H]
    \centering
    \captionsetup{width=\linewidth}
    \caption{Distribution of targets in each comment by the training, development, and test sets}
    \label{tbl_numtarget_distribution}
    \begin{tabular}{|c|rrr|}
    \hline
    \multicolumn{1}{|l|}{}                           & \multicolumn{3}{l|}{\textbf{Number of comments}}                                                             \\ \hline
    \multicolumn{1}{|l|}{\textbf{Number of targets}} & \multicolumn{1}{c|}{\textit{Train}} & \multicolumn{1}{c|}{\textit{Dev}} & \multicolumn{1}{c|}{\textit{Test}} \\ \hline
    0                                                & \multicolumn{1}{r|}{883}            & \multicolumn{1}{r|}{154}          & 225                                \\ \hline
    1                                                & \multicolumn{1}{r|}{3,228}          & \multicolumn{1}{r|}{560}          & 838                                \\ \hline
    2                                                & \multicolumn{1}{r|}{2,586}          & \multicolumn{1}{r|}{437}          & 672                                \\ \hline
    3                                                & \multicolumn{1}{r|}{269}            & \multicolumn{1}{r|}{41}           & 51                                 \\ \hline
    4                                                & \multicolumn{1}{r|}{31}             & \multicolumn{1}{r|}{8}            & 14                                 \\ \hline
    5                                                & \multicolumn{1}{r|}{3}              & \multicolumn{1}{r|}{1}            & 0                                  \\ \hline
    \end{tabular}
\end{table}

Besides, Table \ref{tbl_numtarget_distribution} illustrates the distribution of the number of targets in each comment by the training, development, and test sets. It can be seen that most of the comments in the ViTHSD have one and two targets. The comments that have no target also take up a large proportion. Also, some comments have five targets, but the amount is very few. Moreover, we explore the distribution of comments by each target as shown in Table \ref{tbl_target_distribution}. According to Table \ref{tbl_target_distribution}, most of the comments are concentrated on the \textit{individuals} and the \textit{group} target. The \textit{religion/creed} has the least comments. It can be seen that most hate speech is mentioned about individuals or groups of people rather than about religion or politics.

\begin{table}[H]
    \centering
    \captionsetup{width=\linewidth}
    \caption{Distribution of comments by targets}
    \label{tbl_target_distribution}
    \begin{tabular}{|l|r|r|r|}
    \hline
                   & \multicolumn{1}{l|}{\textbf{Train}} & \multicolumn{1}{l|}{\textbf{Dev}} & \multicolumn{1}{l|}{\textbf{Test}} \\ \hline
    individuals     & 5,480                               & 938                               & 1398                               \\ \hline
    groups         & 2,977                               & 517                               & 769                                \\ \hline
    religion/creed & 24                                  & 8                                 & 6                                  \\ \hline
    race/ethnicity & 502                                 & 74                                & 129                                \\ \hline
    politics       & 363                                 & 57                                & 89                                 \\ \hline
    \end{tabular}
\end{table}

\begin{table}[H]
    \centering
    \caption{The number of levels by each target on the training, development, and test sets of the ViTHSD dataset}
    \label{tbl_target_level}
    \resizebox{\textwidth}{!}{
    \begin{tabular}{|l|rrr|rrr|rrr|}
    \hline
                    & \multicolumn{3}{c|}{\textbf{Train}}                                                                                & \multicolumn{3}{c|}{\textbf{Dev}}                                                                                  & \multicolumn{3}{c|}{\textbf{Test}}                                                                                 \\ \hline
    \textbf{Target} & \multicolumn{1}{l|}{\textit{clean}} & \multicolumn{1}{l|}{\textit{offensive}} & \multicolumn{1}{l|}{\textit{hate}} & \multicolumn{1}{l|}{\textit{clean}} & \multicolumn{1}{l|}{\textit{offensive}} & \multicolumn{1}{l|}{\textit{hate}} & \multicolumn{1}{l|}{\textit{clean}} & \multicolumn{1}{l|}{\textit{offensive}} & \multicolumn{1}{l|}{\textit{hate}} \\ \hline
    individuals      & \multicolumn{1}{r|}{2,480}           & \multicolumn{1}{r|}{1,169}               & 1,831                               & \multicolumn{1}{r|}{454}            & \multicolumn{1}{r|}{189}                & 295                                & \multicolumn{1}{r|}{618}            & \multicolumn{1}{r|}{324}                & 456                                \\ \hline
    groups          & \multicolumn{1}{r|}{1,406}           & \multicolumn{1}{r|}{639}                & 932                                & \multicolumn{1}{r|}{266}            & \multicolumn{1}{r|}{96}                 & 155                                & \multicolumn{1}{r|}{356}            & \multicolumn{1}{r|}{185}                & 228                                \\ \hline
    religion/creed  & \multicolumn{1}{r|}{8}              & \multicolumn{1}{r|}{8}                  & 8                                  & \multicolumn{1}{r|}{6}              & \multicolumn{1}{r|}{1}                  & 1                                  & \multicolumn{1}{r|}{3}              & \multicolumn{1}{r|}{2}                  & 1                                  \\ \hline
    race/ethnicity  & \multicolumn{1}{r|}{120}            & \multicolumn{1}{r|}{163}                & 219                                & \multicolumn{1}{r|}{15}             & \multicolumn{1}{r|}{24}                 & 35                                 & \multicolumn{1}{r|}{40}             & \multicolumn{1}{r|}{34}                 & 55                                 \\ \hline
    politics        & \multicolumn{1}{r|}{37}             & \multicolumn{1}{r|}{81}                 & 245                                & \multicolumn{1}{r|}{4}              & \multicolumn{1}{r|}{11}                 & 42                                 & \multicolumn{1}{r|}{10}             & \multicolumn{1}{r|}{21}                 & 58                                 \\ \hline
    \end{tabular}
    }
\end{table}

Finally, Table \ref{tbl_target_level} illustrates the distribution of comments by level on each target. According to Table \ref{tbl_target_level}, the number of clean comments in the \textit{individuals} and \textit{group} is more than comments in offensive and hate levels. However, the difference between the clean and hate in those targets is not significant. In contrast, for \textit{race/ethnicity} and \textit{politics}, the number of comments in the hate level is dramatically higher than the two remaining levels. This shows the trends in the comments mentioned about politics and ethnicity that those comments often express hate feelings. Table \ref{tab_example_comments} shows several example comments and their labels in the ViTHSD dataset.

\begin{table}[H]
    \centering
    \caption{Example comments from the ViTHSD dataset}
    \label{tab_example_comments}
    \resizebox{\textwidth}{!}{
    \begin{tabular}{|l|p{8.5cm}|l|}
    \hline
    \textbf{\#No} & \textbf{Comments}  & \textbf{Labels}  \\
    \hline
    1    & \textit{có học thức chỉ tiếc văn hoá quá kém ...ý thức thua cả một con chó (English: Educated but has bad behavior. Thinking and acting are worse than a dog)}                                                                                                               & \makecell[{{p{3cm}}}]{\textit{{[}individuals\#hate} \\ \textit{group\#hate{]}}}           \\ \hline
    2             & Đi cách ly chứ đâu phải đi du lịch (English: This is the quarantine, not a tour)                                                                                                                                                                                          & \makecell[{{p{3cm}}}]{\textit{{[}individuals\#clean} \\ \textit{group\#clean{]}}}         \\ \hline
    3             & quan trọng là dân đéo tin vào chúng mày nữa chứ ko phải là sợ chúng mày lạm quyền trưng dụng (English: the important thing is that the people do not believe you instead of afraid your abuse of power for requisition)                                                     & \makecell[{{p{3cm}}}]{\textit{{[}group\#offensive} \\ \textit{politic\#offensive{]}}}    \\ \hline
    4             & Loạn hết cmn rồi . Hazz. (English: It is a f**king chaos. How bad the world is!!)                                                                                                                                                                                          & \makecell[{{p{3cm}}}]{\textit{{[}individuals\#offensive} \\ \textit{group\#offensive{]}}} \\ \hline
    5             & Bán hàng vậy mà cũng có ng mua , vì đẹp bất chấp nhục .Con này về miền bắc VN được đón tiếp nồng hậu lắm đây . (English: she sold like a sh*t, just aimed to interface but does not care about morals, but has many customers. This women will be warmly welcome in the North & \makecell[{{p{3cm}}}]{\textit{{[}individuals\#hate} \\ \textit{race/ethnic\#offensive{]}}} \\ \hline
    \end{tabular}
    }
\end{table}

\section{Methodology}
\label{method}

\subsection{Baseline Models}

\begin{figure}[H]
\centering
  \includegraphics[width=\textwidth]{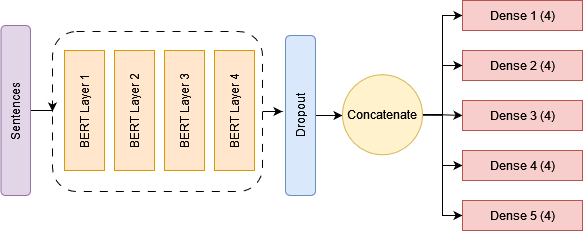}
    \caption{The fine-tunned BERT model for the targeted hate speech detection}
    \label{fig_bert_baseline}
\end{figure}

We follow the system architecture in \cite{9865479} to construct the baseline model for the targeted hate speech detection task. Figure \ref{fig_bert_baseline} illustrates the fine-tuned BERT model for this task. According to the original work \cite{9865479}, we unfreeze the four last layers of BERT for training parameters. Then, the parameters from the BERT are passed to the dropout layer and spread the probability to the five Dense layers. These five Dense layers correspond to five targets in the dataset, including Individual, Group, Religion, Race, and Politics. On each target, there are four levels, including Normal, Clean, Offensive, and Hate (as described in Section \ref{dataset}). We employ the Softmax function to compute the distribution probability to the four levels: normal, clean, offensive, and hate for each Dense layer. The predicted hatred levels corresponding to each target are which have the highest probability from each of the five Dense layers. If the target is predicted as "Normal" level, it means the target is not mentioned in the comment. Otherwise, the remaining levels (Clean, Offensive, and Hate) determine the hatred intensity for each target. To adapt the BERT for the Vietnamese language model, we used the pre-trained multilingual and monolingual BERTology as shown in Table \ref{tab_pretrained_models}.

\begin{table}[ht]
    \centering
    \caption{Pre-trained BERTology models}
    \label{tab_pretrained_models}
    \begin{tabular}{c|c|c}
        \hline
        \textbf{Model} & \textbf{Pre-trained} & \textbf{\# parameters}\\
        \hline
        BERT\cite{devlin-etal-2019-bert} & bert-base-multilingual-cased & 179M \\
        \hline
        XLM-R\cite{conneau-etal-2020-unsupervised} & xlm-roberta-base & 279M \\
        \hline
        PhoBERT\cite{nguyen-tuan-nguyen-2020-phobert} & phobert-base & 135M \\
        \hline
        VELECTRA\cite{bui-etal-2020-improving} & velectra-base-discriminator-cased & 110M \\
        \hline
        ViSoBERT\cite{nguyen2023visobert} & uitnlp/visobert & 97M\\
        \hline
    \end{tabular}
\end{table}

\begin{figure}[H]
\centering
  \includegraphics[width=\textwidth]{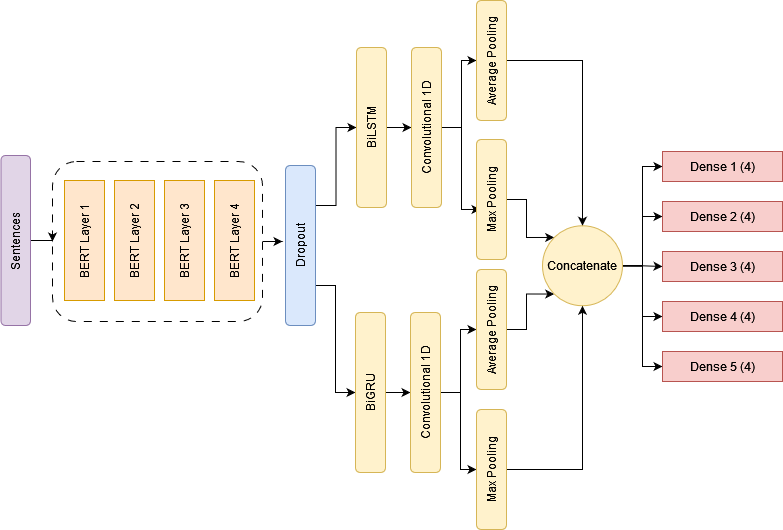}
    \caption{The fine-tunned BERT model combined with the Bi-GRU-LSTM-CNN models}
    \label{fig_bert_bi-gru-lstm-cnn}
\end{figure}

In addition, to enhance the ability of the model to extract valuable features from the texts for hate speech detection, we combine the Bi-GRU-LSTM-CNN \cite{van2019hate} with the pre-trained BERTology. The Bi-GRU-LSTM-CNN \cite{van2019hate} employs 2 robust sequence processing models including Bi-GRU and Bi-LSTM to extract valuable features from the input sequence. Then it uses a CNN layer to fine-grained the information from Bi-GRU and Bi-LSTM before forwarding it to the Dense layer for classification. The Bi-GRU-LSTM-CNN model achieved optimistic results from the VLSP 2019 - Hate speech detection shared task \cite{vu2020hsd}. Figure \ref{fig_bert_bi-gru-lstm-cnn} shows our proposed model architecture for targeted hate speech detection, which is a modification of the original Bi-GRU-LSTM-CNN model \cite{van2019hate}. First, we replace the embedding layer with the pre-trained BERT layers to take advantage of BERT as a robust language model for text representation. Then, for the last layer, we use five Dense layers instead to treat the probability distribution of hatred levels within five targets. 

Finally, for the pre-trained BERTology models as mentioned in Table \ref{tab_pretrained_models}, we choose the model which has the highest performance to construct the prediction model for the online streaming system (as described in Section 4.2). Our empirical results of each model are illustrated in Section \ref{results}.

\subsection{Online Streamming System}

Various real-time data processing methods and systems have been introduced, such as the data streaming process by Nagarajan et al.\cite{nagarajan2019classifying} and the real-time processing of Twitter tweets using Spark Streaming \cite{zaharia2016apache}. Rengarajan et al. \cite{rengarajan2020generalizing} has also presented a streaming pipeline design for big data with Kafka \cite{hiraman2018study}. Studies have been done on real-time streaming systems to detect hate speech in Vietnamese. For example, Long-An Doan et al. \cite{9994299} used machine learning and big data technology to detect hate speech in Vietnamese YouTube comments. Another study, by Khanh Q. Tran et al. \cite{quoc2023vietnamese}, developed a novel hate speech detection model that combines a pre-trained PhoBERT model and a Text-CNN model and uses EDA techniques to improve classification performance. To sum up the paradigm for constructing a system that can process a real-time stream of data, Dinh et al. \cite{10318848} proposed a scalable framework that includes four main components. First is the crawling modules that collect real-time text comments from social networks. Then, these text comments are pre-processed to create clean text. Next, the cleaned data is fitted to the classification model, which has been trained and evaluated before, to make the prediction labels. Finally, to deliver valuable information to the end-users, the prediction results are visualized on a real-time dashboard. 

However, all of these systems are constructed for the multiclass classification, which is inadequate for the task of targeted hate speech detection (as described in Section \ref{intro}). Inheriting previous works on constructing real-time Hate Speech Detection systems, we introduce a system capable of recognizing real-time streaming video comments from social media platforms. It works by processing comments as they are posted and flagging comments that show signs of hate speech. The system detects not only offensive language, but also the objects and aspects that users mention, and the level to which they mention them.

\begin{figure}[ht!]
\centering
  \includegraphics[width=\textwidth]{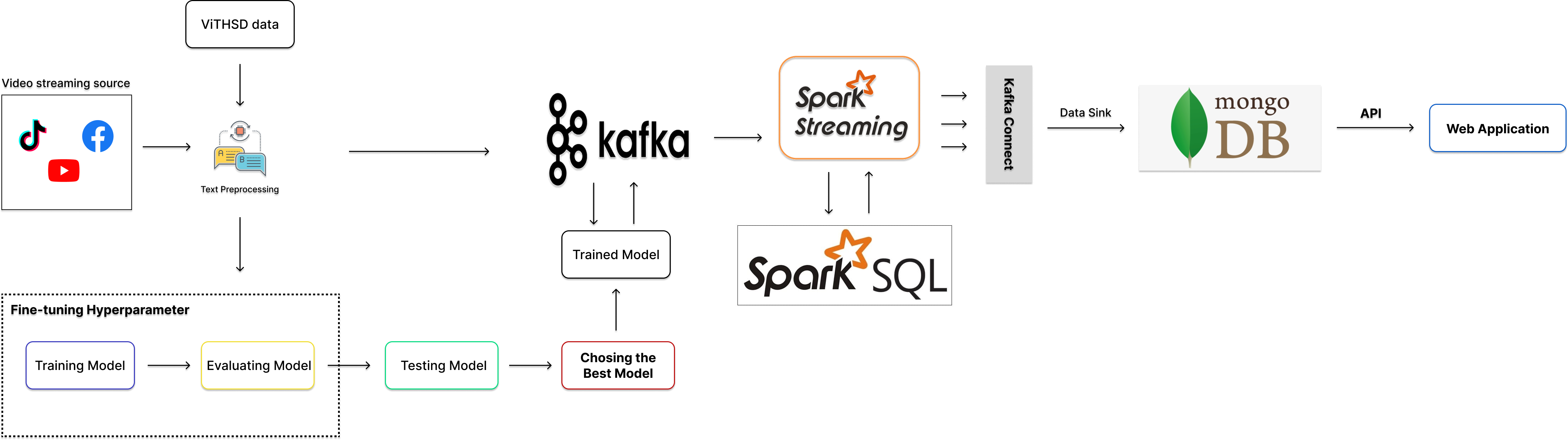}
    \caption{System Architecture for the Detection of Vietnamese Hate Speech from Video Streams on Social Media in Vietnam}
    \label{System Architecture}
\end{figure} 

\textbf{Streaming Processing System}: Data streaming, which is necessary for processing large amounts of live data, involves the continuous collection of real-time data from multiple sources. Social networking platforms such as Facebook, YouTube, and TikTok are some of the most common sources of such data. Real-time data streaming approaches such as Apache Kafka, and Spark Streaming are commonly used for this. We focus on implementing Spark Streaming \cite{zaharia2016apache} for data streaming in this article.

\textbf{Input data}: Our focus is on processing comments from live videos, which requires the system to crawl data based on the video's ID.

\textbf{Main model}: Data pre-processing and hate speech detection on the collected data.

\textbf{Output data}: The list comprises terms in the format $target\#level$, where $target$ refers to the target entity mentioned in the comments, and $level$ indicates the severity level of hate speech detected concerning the target.

\textbf{Data pre-processing}: Fig \ref{System Architecture} offers a system architecture overview for detecting Vietnamese hate speech. Sections \ref{dataset} and \ref{method} discuss the process of building the ViTHSD data set, model training, and criteria used to select the best model. To account for noisy data present in social platform input, especially streaming video, we implement a data preprocessing process to enhance the dataset's quality and extract valuable features. Our process is based on the two-phase data preprocessing method of Khanh Q. Tran et al. \cite{quoc2023vietnamese}.

\textbf{Main process}: Data is first pre-processed and normalized to ensure that it is in a consistent format. This data is then fed into Kafka topics, which are a distributed messaging system that allows for real-time data streaming. The data in the Kafka topics is then consumed by a trained model, which uses it to generate predictions. The predictions are then organized and presented by Spark Streaming, which is a real-time processing framework. SparkSQL queries and displays the data, while Kafka Connect sinks the predictions into MongoDB, a document database. This allows for the predictions to be queried and used to build applications, enabling real-time analytics and data mining.

\section{Performance Results and Discussion}
\label{results}

\subsection{Evaluation Metrics}
The targeted hate speech detection is categorized as a multi-label task, which is similar to the ABSA task. Hence, we refer to the evaluation metrics from the VLSP 2018 shared task about sentiment analysis \cite{Nguyen_Nguyen_Ngo_Vu_Tran_Ngo_Le_2019}, including Precision, Recall, and F1-score. According to \cite{Nguyen_Nguyen_Ngo_Vu_Tran_Ngo_Le_2019}, we measure the performance of models by two phases: the target and the target with polarity. Let P is the predicted target and T be the true target. The precision, recall, and F1-score are computed as shown in Equation \ref{precision}, Equation \ref{recall}, and Equation \ref{f1score} respective.

\begin{equation}
    \label{precision}
    Precision = \frac{|P \cap T|}{|P|} 
\end{equation}

\begin{equation}
    \label{recall}
    Recall = \frac{|P \cap T|}{|T|} 
\end{equation}

\begin{equation}
    \label{f1score}
    F1 = 2*\frac{Precision*Recall}{Precision+Recall} 
\end{equation}

For the target with level evaluation, the P and T are the tuples of the target level instead. The precision, recall, and F1-score are the same as in Equation \ref{precision}, Equation \ref{recall}, and Equation \ref{f1score}.  

\subsection{Empirical Results}
Table \ref{tab_empirical_results} illustrates the results of machine learning models on the ViTHSD test set for the two tasks: target prediction only and the full of target and levels. The empirical results show that the XLM-R model shows better results than other BERTology models and the Bi-GRU-LSTM-CNN model on both target and \textit{target\#level} detections. For target detection, the XLM-R model achieved 71.62\% by F1 score and 50.84\% by F1-score for the target with level detection. Besides, for the monolingual models on the Vietnamese language, the ViSoBERT achieved the best results on both target and \textit{target\#level} detection, which are 70.10\% for target detection and 51.12\% for the \textit{target\#level} detection. It can be seen that for the \textit{target\#level} task, ViSoBERT is better than XLM-R, while its result is slightly lower than XLM-R on the target detection task. 

On the other hand, the BERTology models show robustness performance in comparison with the original Bi-GRU-LSTM-CNN model on both tasks. Instead, the performance of the XLM-R - Bi-GRU-LSTM-CNN is slightly higher than the XLM-R and ViSoBERT, which is 72.01\% by F1 score for the target prediction and 51.74\% by F1 score for the \textit{target\#level} task. The ViSoBERT when combined with Bi-GRU-LSTM-CNN (ViSoBERT - Bi-GRU-LSTM-CNN) does not obtain the improvement in comparison with the ViSoBERT only. However, the ViSoBERT - Bi-GRU-LSTM-CNN achieved higher results than XLM-R - Bi-GRU-LSTM-CNN on the \textit{target\#level} detection task by Recall metric.

\begin{table}[ht]
    \caption{Empirical results of baseline models on the dataset}
    \label{tab_empirical_results}
    \begin{tabular}{|p{2.5cm}|lll|lll|}
    \hline
    \textbf{Model}                   & \multicolumn{1}{l|}{\textbf{F1 score}} & \multicolumn{1}{l|}{\textbf{Precision}} & \textbf{Recall} & \multicolumn{1}{l|}{\textbf{F1 score}} & \multicolumn{1}{l|}{\textbf{Precision}} & \textbf{Recall} \\ \hline
                                     & \multicolumn{3}{c|}{\textit{\textbf{Target only}}}                                                 & \multicolumn{3}{c|}{\textit{\textbf{Target + level}}}                                           \\ \hline
    Bi-GRU-LSTM-CNN                  & \multicolumn{1}{l|}{65.73}             & \multicolumn{1}{l|}{67.18}              & 64.33           & \multicolumn{1}{l|}{46.65}             & \multicolumn{1}{l|}{47.66}              & 45.68           \\ \hline
    m-BERT                           & \multicolumn{1}{l|}{67.83}             & \multicolumn{1}{l|}{68.51}              & 67.15           & \multicolumn{1}{l|}{47.15}             & \multicolumn{1}{l|}{47.26}              & 47.03           \\ \hline
    \textbf{XLM-R}                   & \multicolumn{1}{l|}{\textbf{71.62}}    & \multicolumn{1}{l|}{\textbf{74.50}}     & \textbf{68.96}  & \multicolumn{1}{l|}{\textbf{50.84}}    & \multicolumn{1}{l|}{\textbf{52.33}}     & \textbf{49.44}  \\ \hline
    PhoBERT                          & \multicolumn{1}{l|}{69.87}             & \multicolumn{1}{l|}{73.42}              & 66.65           & \multicolumn{1}{l|}{50.33}             & \multicolumn{1}{l|}{52.24}              & 48.55           \\ \hline
    VELECTRA                         & \multicolumn{1}{l|}{67.87}             & \multicolumn{1}{l|}{70.84}              & 65.13           & \multicolumn{1}{l|}{46.58}             & \multicolumn{1}{l|}{48.20}              & 45.06           \\ \hline
    \textbf{ViSoBERT}                         & \multicolumn{1}{l|}{70.10}             & \multicolumn{1}{l|}{71.53}              & 68.73           & \multicolumn{1}{l|}{\textbf{51.12}}             & \multicolumn{1}{l|}{51.77}              & \textbf{50.49}           \\ \hline
    \textbf{XLM-R - Bi-GRU-LSTM-CNN} & \multicolumn{1}{l|}{\textbf{72.01}}    & \multicolumn{1}{l|}{\textbf{75.26}}     & \textbf{69.02}  & \multicolumn{1}{l|}{\textbf{51.74}}    & \multicolumn{1}{l|}{\textbf{53.68}}     & \textbf{49.93}  \\ \hline
    \textbf{ViSoBERT - Bi-GRU-LSTM-CNN} & \multicolumn{1}{l|}{69.49}    & \multicolumn{1}{l|}{71.37}     & 67.70  & \multicolumn{1}{l|}{51.16}    & \multicolumn{1}{l|}{52.23}     & \textbf{50.14}  \\ \hline
    \end{tabular}
\end{table}

In conclusion, according to Table \ref{tab_empirical_results}, the BERTology models show the robust ability for the target hate speech detection task. The XLM-R, which is a multilingual pre-trained model, obtained the best results on the ViTHSD dataset for the target detection task. In contrast, the ViSoBERT, which is a monolingual pre-trained model on Vietnamese social media texts, achieved the best results on the ViTHSD dataset for the \textit{target\#level} detection task. Besides, our proposed model, which combined the XLM-R with the Bi-GRU-LSTM-CNN, showed the best result on the ViTHSD on both tasks. However, from the empirical results in Table \ref{tab_empirical_results}, there is a large gap between the target detection and the target with level detection task, indicating the challenge of constructing models for the target with level prediction task on the ViTHSD dataset. Compared to two monolingual pre-trained models for Vietnamese texts including PhoBERT and VELECTRA, the ViSoBERT performance on the ViTHSD dataset is better than the two models. 

\subsection{Error Analysis and Discussion}

\begin{figure}[!h]
  \centering
  \begin{subfigure}[b]{0.45\linewidth}
    \includegraphics[width=\linewidth]{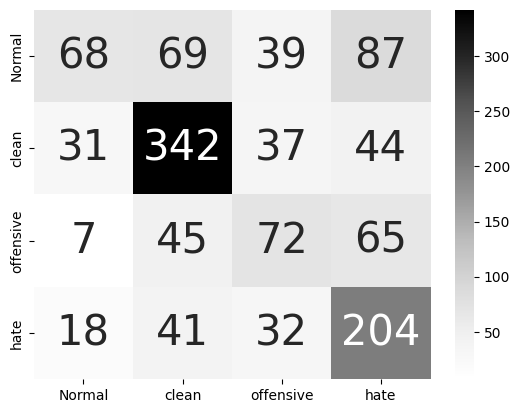}
    \caption{Individual.}
    \label{fig:Individual}
  \end{subfigure}
  \hfill
  \begin{subfigure}[b]{0.45\linewidth} \includegraphics[width=\linewidth]{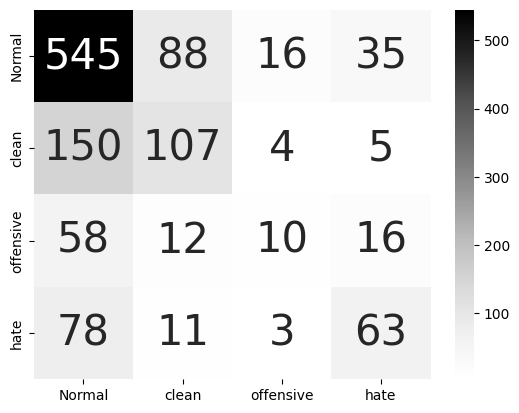}
    \caption{Groups.}
    \label{fig:groups}
  \end{subfigure}
    \hfill
  \begin{subfigure}[b]{0.45\linewidth} \includegraphics[width=\linewidth]{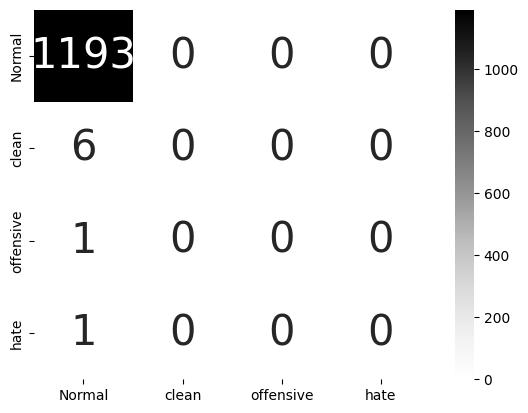}
    \caption{religion/creed .}
    \label{fig:religion}
  \end{subfigure}
    \hfill
  \begin{subfigure}[b]{0.45\linewidth} \includegraphics[width=\linewidth]{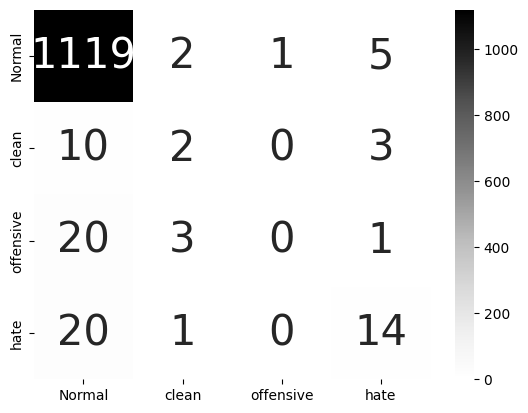}
    \caption{race/ethnicity.}
    \label{fig:race}
  \end{subfigure}
    \hfill
  \begin{subfigure}[b]{0.45\linewidth} \includegraphics[width=\linewidth]{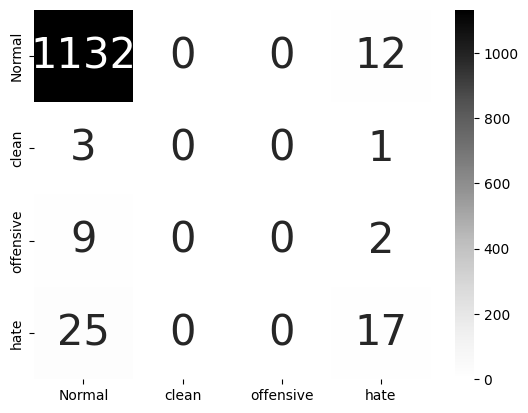}
    \caption{Politics.}
    \label{fig:politics}
  \end{subfigure}
  
  \caption{Confusion matrix of XLM-R - Bi-GRU-LSTM-CNN model on development dataset of the ViTHSD dataset on each target. The Normal label indicates that the target has no mention.}
  \label{fig:hinh_group}
\end{figure}

Figure \ref{fig:hinh_group} displays the Confusion Matrix of the XLM-R - Bi-GRU-LSTM-CNN model on the development dataset of the ViTHSD dataset. As for individual aspects of the dataset, the model predicts that most comments labeled as hateful are actually just normal comments. Moving on to the group aspect, normal comments are wrongly classified by the classifier as either clean or hateful. In terms of other aspects of the dataset, the majority of labels were accurately predicted, despite some normal comments being misclassified as either hateful or offensive. Overall, it can be seen that the imbalance in the dataset between targets affects the performance of the classification model. For the religion target, the results completely fell into the class normal, and no correct results for the clean, offensive, and hate labels. Hence, this shows that the model could not give the correct prediction for this target because of the limitation of data for training (see Table \ref{tbl_target_level}). Besides, with the race/ethnicity target, the classification model tends to give the prediction of clean and hate although the number of correct predictions is minor compared to the normal labels. Also in this target, the number of correct hate predictions is larger than clean although it does not give any correct prediction for the offensive label. For the political target, it is shown that most of the correct predictions are in the hate label and no correct prediction on clean and offensive. This indicates the trends in the classification model that it mostly returns hate labels when the target race/ethnicity and politics are detected. 

\begin{table}[ht]
    \caption{Wrong prediction samples on the development dataset of the ViTHSD dataset}
    \label{prediction_samples}
    \centering
    \resizebox{\textwidth}{!}{  
    \begin{tabular}{|c|c|c|c|}
    \hline
    \textbf{\#} & \textbf{Comments} & \textbf{Predict } & \textbf{True} \\ 
    \hline
    1 & \multicolumn{1}{m{5cm}|} {Dm \newline
    (English: F*ck you)}  & \multicolumn{1}{m{3cm}|} {individual\#\textbf{offensive} \newline politics\#\textbf{hate}}  & \multicolumn{1}{m{3.3cm}|} {individual\#\textbf{offensive} \newline groups\#\textbf{offensive} } \\
    \hline
    2 & \multicolumn{1}{m{5cm}|} {Như cl luôn =)) \newline
    (English: Same as c*nt =))}  & \multicolumn{1}{m{3cm}|} {individual\#\textbf{normal} \newline groups\#\textbf{offensive}}  & \multicolumn{1}{m{3.3cm}|} {individual\#\textbf{offensive} \newline groups\#\textbf{offensive} } \\
    \hline
     3 & \multicolumn{1}{m{5cm}|} {Hành kinh < tên quốc gia> đi, chung lòng cứu quốc... \newline
    (English: Soldiers of <country name's>, marching onward
    United in determination to save the nation)}  & \multicolumn{1}{m{3cm}|} {race/ethnicity\#\textbf{clean}}  & \multicolumn{1}{m{3.3cm}|} {race/ethnicity\#\textbf{hate}}\\
    \hline
    4 & \multicolumn{1}{m{5cm}|} {Xin hỏi nhà tiên tri.sau đợt dịch này bọn 3/// liệu có bị tiệt chủng không \newline
    (English: Esteemed Prophet, after this pandemic, is it possible that 3/// will meet its end?)}  & \multicolumn{1}{m{3cm}|} {individual\#\textbf{clean} \newline religion/creed\#\textbf{normal} }  & \multicolumn{1}{m{3.3cm}|} {individual\#\textbf{hate} \newline religion/creed\#\textbf{offensive} } \\
    \hline
    5 & \multicolumn{1}{m{5cm}|} {Giáo án cc gì đây \newline
    (English: What is the cc lesson plan?)}  & \multicolumn{1}{m{3cm}|} {groups\#\textbf{normal}}  & \multicolumn{1}{m{3.3cm}|} {groups\#\textbf{hate}} \\
    \hline
    \end{tabular}
    }
\end{table}

In addition, Table \ref{prediction_samples} shows incorrect predictions by the  XLM-R - Bi-GRU-LSTM-CNN cased model. Comments 1, 2, and 5 include slang and abbreviations frequently used by Vietnamese social media users, like "Dm," "cl," and "cc." What's peculiar here is that when these words stand alone, they express a negative sentiment, as in comment 1, where "Dm" is particularly offensive. However, the level of offense varies depending on the context in which they are used, as in comments 2 and 5. These special words confused the model and misclassified them from offensive or hateful to normal. Additionally, comments 3, and 4 have acronyms and words with covert connotations. "Hành kinh" and "3///" have ambiguous interpretations leading to the model's erroneous classification. From the five samples, the incorrect predictions are due to the lexicon problem in social media text where the language is diverse and acronyms, slang, and code-mixing are used by social media users. Therefore, it is necessary to have a lexicon normalization \cite{MEHMOOD2020102368} application in social media texts for pre-processing before fitting into the model. Furthermore, from the error analysis, we found the difficulty of classification comes from the deep meaning hidden between text about the target and holder entities mentioned in the sentence. This is the essential feature to detect the correct target in the hateful comments, which is currently the drawback of our model on the minority target due to the imbalance in the dataset. One of the potential approaches to improve the identification of the target in the sentence is the othering language embedding \cite{10.1145/3324997}, which computes the semantic distances of the part-of-speech (POS) features in the sentence to retrieve the correct target of hatred language. Finally, we hope that the large language modeling will help to solve the problem of deep contextual and implicit reasoning in the hateful text when it is pre-trained on social media texts for Vietnamese. This will help to improve the accuracy of the targeted hate speech detection task. 

\section{Online Streaming Results}
\label{result_online_stream}
\subsection{Execution Evaluation on Streaming Data}
\begin{figure}[ht]
\centering
  \includegraphics[width=\textwidth]{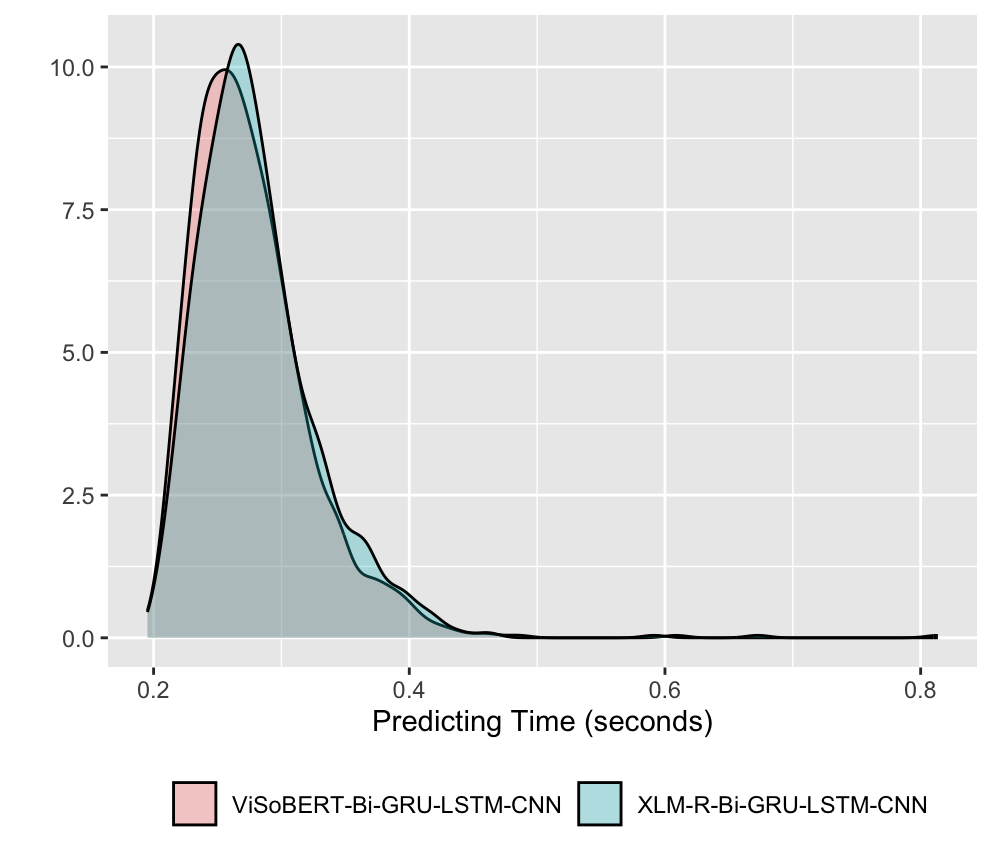}
    \caption{Distribution of prediction time in the streaming system of XLM-R-Bi-GRU-LSTM-CNN and ViSoBERT-Bi-GRU-LSTM-CNN}
    \label{fig_prediction_distribution}
\end{figure}

We investigate the performance of XLM-R-Bi-GRU-LSTM-CNN and the ViSoBERT-R-Bi-GRU-LSTM-CNN on a streaming system to measure the efficiency of both models in a practical real-time application. We integrate two models into the streaming systems to predict about 1,000 comments from YouTube live streaming in real time. Then we measure the time execution (milliseconds) for predicting each comment of each model and summarize them in quartile information. Table \ref{tab_stat_time} shows the statistical of time execution for each model, and Figure \ref{fig_prediction_distribution} illustrates the distribution of time execution for each comment of both XLM-R-Bi-GRU-LSTM-CNN and ViSoBERT-R-Bi-GRU-LSTM-CNN models. We run the experiments on the server with 16GB RAM, 50GB Disk SSD SATA 3, and 4 CPU Intel Core i5 (i5-8250U).

\begin{table}[ht]
    \centering
    \captionsetup{width=\linewidth}
    \caption{Statistical information about prediction times (milliseconds) and size of classification models (Gigabyte) on the streaming system.}
    \label{tab_stat_time}
    \begin{tabular}{lrrrrrr|r}
            \multicolumn{1}{l}{\textbf{Model}} & \multicolumn{1}{l}{\textbf{Min}} & \multicolumn{1}{l}{\textbf{25\%}} & \multicolumn{1}{l}{\textbf{Median}} & \multicolumn{1}{l}{\textbf{Mean}} & \multicolumn{1}{l}{\textbf{75\%}} & \multicolumn{1}{l}{\textbf{Max}} & \multicolumn{1}{l}{\textbf{Size}} \\
             \hline
        XLM-R & 173.9 & 216.1 & 246.3 & 251.2 & 277.8 & 465.6 & 3.11  \\
        ViSoBERT & 164.6 & 209.5 & 233.5 & 239.6 & 260.6 & 701.9 & 1.09 \\  
        XLM-R-Bi-GRU-LSTM-CNN    & 201.2                            & 249.1                                     & 272.6                                & 281.4                             & 303.2                                     & 672.6           & 3.18                 \\
        ViSoBERT-Bi-GRU-LSTM-CNN & 195.2                            & 243.6                                     & 268.3                               & 276.7                            & 298.7                                     & 813.5       & 1.16                    \\
        \hline
    \end{tabular}
\end{table}

According to Table \ref{tab_stat_time}, it can be seen that the execution time of ViSoBERT and ViSoBERT-Bi-GRU-LSTM-CNN are significantly less than XLM-R and XLM-R-Bi-GRU-LSTM-CNN for about 10ms on average. Besides, the size of ViSoBERT is significantly less than XLM-R because has smaller parameters than XLM-R (as mentioned in Table \ref{tab_pretrained_models}). Also, the size of ViSoBERT-Bi-GRU-LSTM-CNN is lighter than XLM-R-Bi-GRU-LSTM-CNN.

Additionally, as shown in Figure \ref{fig_prediction_distribution}, most of the comments predicted by the ViSoBERT-Bi-GRU-LSTM-CNN are faster than the XLM-R-Bi-GRU-LSTM-CNN. This is meaningful in the real-time prediction in online streaming systems, which help process a large amount of comments in a timely manner. 

Finally, ViSoBERT seems to be the best choice for constructing the real-time targeted hate speech detection system for online streaming comments since it has a performance similar to the XLM-R for Vietnamese text with lighter parameters and faster processing. 

\subsection{Targeted Hate Speech Detection with Streaming Data}
\begin{figure}[ht!]
    \centering
    \includegraphics[width=.8\textwidth]{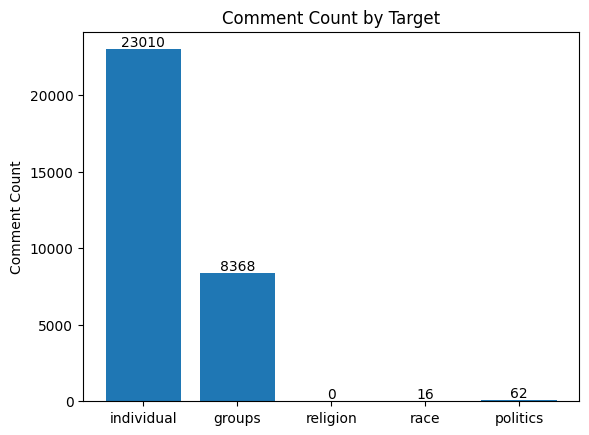}
    \caption{Total comments contain hate speech types from our online detection system. }
    \label{fig_comment_count}
\end{figure}

\begin{figure}[ht!]
  \centering
  \begin{subfigure}[b]{0.45\linewidth}
    \includegraphics[width=\linewidth]{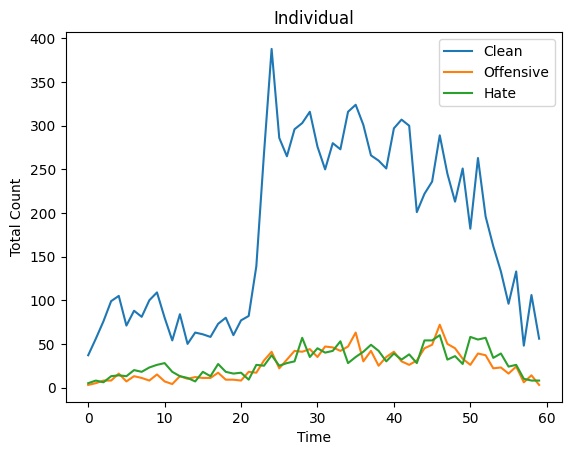}
    \caption{Individual.}
  \end{subfigure}
  \hfill
  \begin{subfigure}[b]{0.45\linewidth} \includegraphics[width=\linewidth]{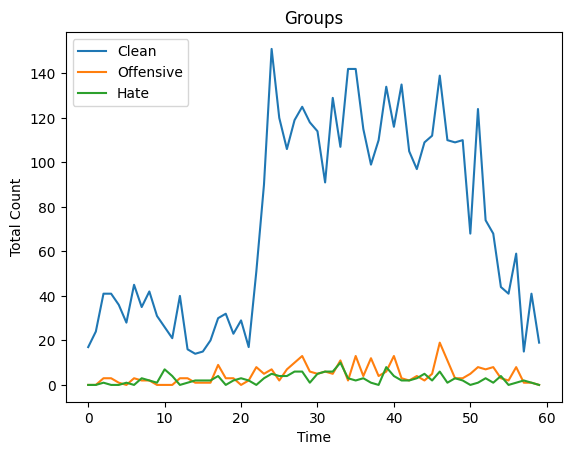}
    \caption{Groups.}
  \end{subfigure}
    \hfill
  \begin{subfigure}[b]{0.45\linewidth} \includegraphics[width=\linewidth]{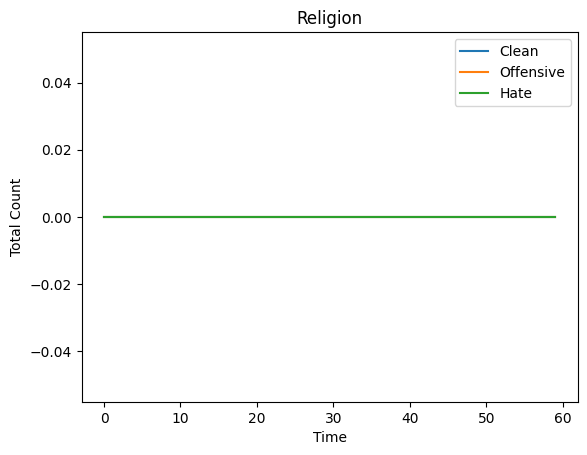}
    \caption{religion/creed .}
  \end{subfigure}
    \hfill
  \begin{subfigure}[b]{0.45\linewidth} \includegraphics[width=\linewidth]{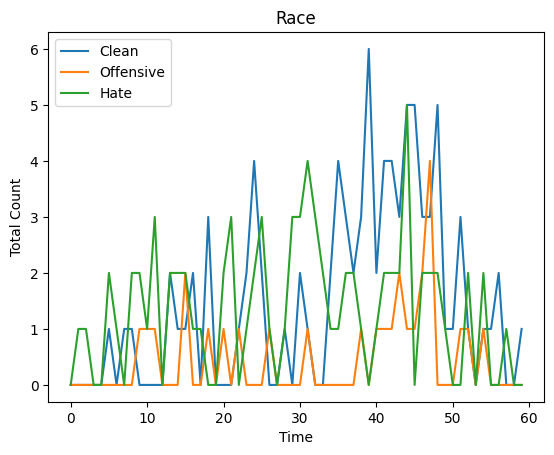}
    \caption{race/ethnicity.}
  \end{subfigure}
    \hfill
  \begin{subfigure}[b]{0.45\linewidth} \includegraphics[width=\linewidth]{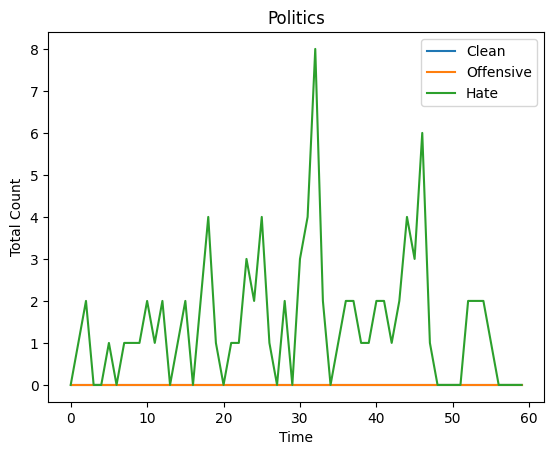}
    \caption{Politics.}
  \end{subfigure}
  \caption{Results of detected targets by our online detection system.}
  \label{fig:online_streaming_results}
\end{figure}

We use our online detecting system to detect harmful comments from streaming videos on YouTube in 60 minutes. We collected a total of 15,307 comments during the streaming video. Then we statistic the number of detected labels of each target by one minute. According to Figure \ref{fig_comment_count}, most predicted comments mentioned individual and group targets. The number of comments aiming to \textit{race/ethnicity} and \textit{politics} targets are very few. Specifically, no comment about the \textit{region} target is mentioned. Besides, the number of clean comments in the \textit{individual} and \textit{groups} is more than two remains, according to Figure \ref{fig:online_streaming_results}. In general, the comments that are predicted as offensive and hateful do not change too much as the clean comments during the video streaming. In contrast, for the \textit{race/ethnicity} target, the clean, offensive, and hate labels change dramatically in 60 minutes of streaming video. For the \textit{politics} target, the system seems to predict comments in the streaming video as hate.

Overall, from these preliminary results, we would like to show the application of our system in practice. However, due to the limitation of time as well as the computing resources, we could not test our system on other platforms at the same time. We hope our current results are the case study for the application of the proposed streaming system to detect harmful comments by targets on online social platforms.

\section{Conclusion and Future Work}
\label{conclusion}
In this paper, we constructed a new dataset called ViTHSD to detect targeted hate speech in Vietnamese texts. The dataset comprises 10,000 comments manually annotated by humans across five targets: Individual (1), Group (2), Religion/creed (3), Race/Ethnicity (4), and Politics (5). Each target includes four levels of hate: NORMAL, CLEAN, OFFENSIVE, and HATE, except NORMAL which indicates no specific target. We implemented a comprehensive annotation process, from annotator selection to consensus evaluation methodology. The ViTHSD dataset has a moderate level of inter-annotator agreement, measured by the Cohen-Kappa K-average metric ($k$=0.45). We assess multiple aspects of hate speech detection by testing several state-of-the-art models on this dataset. The XLM-R Bi-GRU-LSTM-CNN model achieves the best performance, with a Macro F1 score of 72.01\% for the Target task only and 51.74\% for the Target task with a level. Besides, the ViSoBERT - a monolingual pre-trained model on Vietnamese social texts, obtained better performance than XLM-R with fewer parameters and less time execution. Additionally, we have successfully constructed a streaming architecture that runs smoothly and demonstrated its functionality through a continuous 60-minute video stream, indicating the potential application of our work in the practical intelligence social listening systems to reduce hatred content. 

Challenges such as vocabulary inconsistencies in social media texts, which include diverse languages, acronyms, slang, and mixed codes, result in inaccurate predictions. Addressing low-accuracy models will be a significant challenge going forward. Our error analysis revealed that developing a system capable of recognizing offensive language on social networks in Vietnam is incredibly challenging due to the complexity of the social texts created by users on social networks. Hence, our next studies will focus on improving the ability to understand the social texts of the model by normalizing the acronyms, slang, and code-mixing. Furthermore, we leverage the robust ability of the Large-language models (LLMs) to improve the performance of targeted hate speech detection tasks in the Vietnamese language.


\section*{Statements and Declarations}
\subsection*{Availability of supporting data}
The source code and data are available for research purposes at: \url{https://github.com/bakansm/ViTHSD}

\subsection*{Competing interests}
The authors declare that they have no competing interests. 

\subsection*{Funding}
This research was supported by The VNUHCM-University of Information Technology's Scientific Research Support Fund.

\bibliography{references}

\end{document}